\documentclass[journal]{IEEEtran}
\usepackage{hyperref,tablefootnote,footnotehyper}
\usepackage{multirow}
\usepackage{times}
\usepackage{epsfig}
\usepackage{graphicx}
\usepackage{amsmath}
\usepackage{amssymb}
\usepackage{subfigure}
\usepackage{colortbl}
\usepackage{multirow}
\usepackage{soul}
\usepackage{adjustbox}
\usepackage{array}
\usepackage{setspace}
\usepackage{lscape}
\usepackage{mathtools}
\usepackage{adjustbox}
\usepackage{boldline}
\usepackage{pifont}

\usepackage{tikz}
\usetikzlibrary{positioning}
\usetikzlibrary{shapes.geometric}

\definecolor{mblue}{RGB}{0,114,189}
\definecolor{mgreen}{RGB}{91,191,33}
\definecolor{morange}{RGB}{237,177,32}
\definecolor{darkblue}{RGB}{68,49,141}
\definecolor{darkorange}{RGB}{242,184,23}
\newcommand{\printfnsymbol}[1]{%
  \textsuperscript{*}%
} 

\usepackage{array}
\newcolumntype{L}[1]{>{\raggedright\let\newline\\\arraybackslash\hspace{0pt}}m{#1}}
\newcolumntype{C}[1]{>{\centering\let\newline\\\arraybackslash\hspace{0pt}}m{#1}}
\newcolumntype{R}[1]{>{\raggedleft\let\newline\\\arraybackslash\hspace{0pt}}m{#1}}

\begin{document}
%
\title{GANprintR: Improved Fakes and Evaluation of the State of the Art in Face Manipulation Detection}
%
%
%

%

\author{Jo\~{a}o C. Neves\printfnsymbol{1}\thanks{* J.C. Neves and R. Tolosana contributed equally to
this study.}, Ruben Tolosana\printfnsymbol{1}, Ruben Vera-Rodriguez, Vasco Lopes,  Hugo Proen\c{c}a and Julian Fierrez \\
\thanks{J.C. Neves is with NOVA-LINCS, Portugal (e-mail: jcneves@di.ubi.pt)}
\thanks{R. Tolosana, R. Vera-Rodriguez and J. Fierrez are with the Biometrics and Data Pattern Analytics - BiDA Lab, Universidad Autonoma de Madrid, Spain (e-mail: \{ruben.tolosana, ruben.vera, julian.fierrez\}@uam.es).}
\thanks{V. Lopes is with University of Beira Interior, Portugal (e-mail: vasco.lopes@ubi.pt)}
\thanks{H. Proen\c{c}a is with IT - Instituto de Telecomunica\c{c}\~{o}es (e-mail: hugomcp@di.ubi.pt).}}

%



\maketitle

\begin{abstract}
The availability of large-scale facial databases, together with the remarkable progresses of deep learning technologies, in particular Generative Adversarial Networks (GANs), have led to the generation of extremely realistic fake facial content, raising obvious concerns about the potential for misuse. Such concerns have fostered the research on manipulation detection methods that, contrary to humans, have already achieved astonishing results in various scenarios. In this study, we focus on the synthesis of entire facial images, which is a specific type of facial manipulation. The main contributions of this study are four-fold: \textit{i)} a novel strategy to remove GAN ``fingerprints" from synthetic fake images based on autoencoders is described, in order to spoof facial manipulation detection systems while keeping the visual quality of the resulting images; \textit{ii)} an in-depth analysis of the recent literature in facial manipulation detection; \textit{iii)} a complete experimental assessment of this type of facial manipulation, considering the state-of-the-art fake detection systems (based on holistic deep networks, steganalysis, and local artifacts), remarking how challenging is this task in unconstrained scenarios; and finally \textit{iv)} we announce a novel public database, named iFakeFaceDB, yielding from the application of our proposed GAN-fingerprint Removal  approach (GANprintR) to already very realistic synthetic fake images. 

The  results obtained in our empirical evaluation show that additional efforts are required to develop robust facial manipulation detection systems against unseen conditions and spoof techniques, such as the one proposed in this study. 
\end{abstract}


\IEEEpeerreviewmaketitle

\begin{figure}[t]
\begin{center}
   \includegraphics[width=0.89\linewidth]{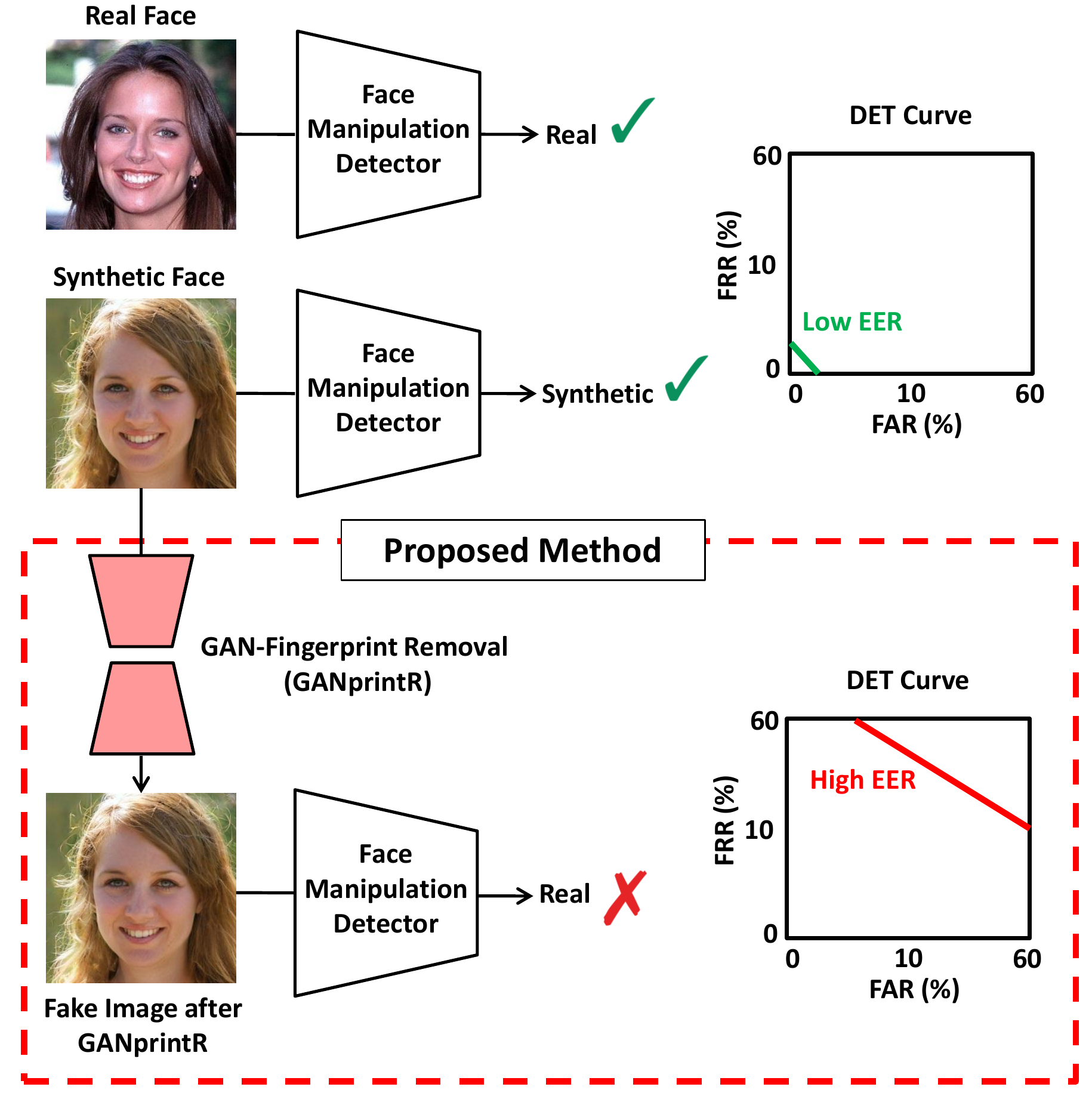}
\end{center}
   \caption{\textbf{Architecture of our proposed GAN-fingerprint removal approach.} In general, state-of-the-art face manipulation detectors can easily distinguish between real and synthetic fake images. This usually happens due to the existence and exploitation by those detectors of GAN ``fingerprints" produced during the generation of synthetic images. We propose an autoencoder module (GANprintR) to remove the GAN fingerprints from the synthetic images and spoof the facial manipulation detection systems, while keeping the visual quality of the resulting images.}
\label{fig:proposed_approach}
\end{figure}

\section{Introduction}
\IEEEPARstart{I}{mages} and videos containing fake facial information obtained by digital manipulation have recently become a great public concern~\cite{BBCNews_deepfake}. So far, the number and realism of digitally manipulated fake facial contents have been limited by the lack of sophisticated editing tools, the high domain of expertise required, and the complex and time-consuming process involved to generate realistic fakes. On the other hand, the scientific communities of biometrics and security in the past decade have been paying growing attention to understanding and protecting against what was considered a relevant threat around face biometrics~\cite{2015_ISPM_PAs}: presentation attacks conducted physically against the face sensor (camera) using various kinds of face spoofs (e.g., 2D or 3D printed, displayed, mask-based, etc.)~\cite{Hernandez-Ortega2019,Galbally_Access_2014}. 

However, nowadays it is becoming increasingly easy to automatically synthesise non-existent faces or even to manipulate the face of a real person in an image/video, thanks to the free access to large public databases and also to the advances on deep learning techniques that eliminate the requirements of manual editing. As a result, accessible open software and mobile applications such as \textit{ZAO} and \textit{FaceApp} have led to large amounts of synthetically generated fake content~\cite{zao,faceapp}.

The most popular methods to generate fake face content can be categorised into four groups, regarding the level of manipulation~\cite{Tolosana_2020_ARXIV,Verdoliva_2020_arxiv, Jain2019facialManipulation}: \textit{i)} entire face synthesis, \textit{ii)} identity swap, \textit{iii)} attribute manipulation, and \textit{iv)} expression swap. 

\begin{table*}[t]
\centering
\caption{Comparison of state-of-the-art manipulation detection approaches for entire face synthesis manipulation.}
\label{table:relatedWorks}
\scalebox{0.78}{
\begin{tabular}{ccccc}
\textbf{Study}                                                                     & \textbf{Features}                                                       & \textbf{Classifiers}      & \textbf{Best Performance} & \textbf{Databases}                                                                \\ \hline
\begin{tabular}[c]{@{}c@{}}McCloskey and Albright (2018)\\ \cite{mccloskey2018detecting}\end{tabular} & Color-related Features                                                  & SVM                       & AUC = 70\%                & NIST MFC2018                                                                      \\ \hline
\begin{tabular}[c]{@{}c@{}}Yu \textit{et al.} (2018)\\ \cite{yu2018attributing}\end{tabular}             & GAN-related Features                                                    & CNN                       & Acc. = 99.50\%            & \begin{tabular}[c]{@{}c@{}}Real: CelebA\\ Fake: Own Database\end{tabular}         \\ \hline

\begin{tabular}[c]{@{}c@{}}Marra \textit{et al.} (2018)\\ \cite{Marra_MIPR_2018}\end{tabular}     & Image-related Features                                                  & CNN & Acc. = 95.07\%  & \begin{tabular}[c]{@{}c@{}}Real: Own Database(CycleGAN)\\ Fake: Own Database(CycleGAN)\end{tabular}    \\ \hline

\begin{tabular}[c]{@{}c@{}}Wang \textit{et al.} (2019)\\ \cite{wang2019fakespotter}\end{tabular}          & \begin{tabular}[c]{@{}c@{}}CNN Neuron \\ Behavior Features\end{tabular} & SVM                       & Acc. = 84.78\%            & \begin{tabular}[c]{@{}c@{}}Real: CelebA-HQ/FFHQ\\ Fake: Own Database\end{tabular} \\ \hline

\begin{tabular}[c]{@{}c@{}}Stehouwer \textit{et al.} (2019)\\ \cite{Jain2019facialManipulation}\end{tabular}     & Image-related Features                                                  & CNN + Attention Mechanism & EER = 0.05\%               & \begin{tabular}[c]{@{}c@{}}Real: CelebA/FFHQ/FaceForensics++\\ Fake: Own Database\end{tabular}    \\ 
\hline

\begin{tabular}[c]{@{}c@{}}Yang \textit{et al.} (2019)\\ \cite{Yang_2019_ICASSP}\end{tabular}     & Head Pose & SVM & AUC = 89\%  & \begin{tabular}[c]{@{}l@{}}Real: UADFV/DARPA MediFor\\ Fake: UADFV/DARPA MediFor\end{tabular}         \\ \hline

\begin{tabular}[c]{@{}c@{}}Matern \textit{et al.} (2019)\\ \cite{Matern_2019_WACVW} \end{tabular} & Eye Color Features & K-NN & AUC = 85.2\%                                      & \begin{tabular}[c]{@{}c@{}}Real: CelebA\\ Fake: Own Database (PGGAN)\end{tabular}     \\ \hline

\begin{tabular}[c]{@{}c@{}}He \textit{et al.} (2019)\\ \cite{He_2019_ICIP}\end{tabular}  & Color-related Features                                & Random Forest & Acc. = 99\% & \begin{tabular}[c]{@{}c@{}}Real: CelebA\\ Fake: Own Database (PGGAN)\end{tabular} \\ \hline

\begin{tabular}[c]{@{}c@{}}Wang \textit{et al.} (2019)\\ \cite{wang2019detecting}\end{tabular}           & Image-related Features                                                  & DRN & AP = 99.8\%             & \begin{tabular}[c]{@{}c@{}}Real: Own Database\\ Fake: Own Database\end{tabular}   \\ \hline
\end{tabular}

}
\end{table*}

In this study, we focus on the entire face synthesis manipulation, where a machine learning model, typically based on Generative Adversarial Networks (GANs)~\cite{goodfellow2014generative}, learns the distribution of the human face data, allowing to generate non-existent faces by sampling this distribution. This type of facial manipulation provides astonishing results, and is able to generate extremely realistic fakes. Nevertheless, contrary to humans, most state-of-the-art detection systems provide very good results against this type of facial manipulation, remarking how easy it is to detect the GAN ``fingerprints" present in the synthetic images. 

In this context, the main contributions of our paper are:

\begin{itemize}
\item A novel approach to spoof state-of-the-art facial manipulation detection systems, while keeping the visual quality of the resulting images. Fig.~\ref{fig:proposed_approach} graphically summarises our proposed approach based on a GAN-fingerprint Removal autoencoder (GANprintR).

\item An in-depth literature analysis of the state-of-the-art detection approaches for the entire face synthesis manipulation, including the key aspects of the detection systems, the databases used for developing and evaluating these systems, and the main results achieved by them.

\item A thorough experimental assessment of this type of facial manipulation considering fake detection (based on holistic deep networks, steganalysis, and local artifacts) and realistic GAN-generated fakes (with and without the proposed GANprintR) over different experimental conditions, i.e., controlled and in-the-wild scenarios. 

\item We announce a novel database named iFakeFaceDB\footnote{\url{https://github.com/socialabubi/iFakeFaceDB}}, resulting from the application of our GANprintR to already very realistic synthetic images.\\

\end{itemize}

The remainder of the paper is organised as follows. Sec.~\ref{related_works} summarises previous studies focused on the detection of the entire face synthesis manipulation. Sec.~\ref{proposed_Approach} explains our proposed GAN-fingerprint removal approach. Sec.~\ref{sec:databases} summarises the key features of the real and fake databases considered in our experimental framework. Sec.~\ref{experimental_protocol} and \ref{experimental_results} describe the proposed experimental setup and results achieved, respectively. Finally, Sec.~\ref{conclusions} draws the final conclusions and points out some lines for future work.


\begin{figure*}[t]
	\centering
	\resizebox{0.85\textwidth}{!}{
	\begin{tikzpicture}
		
		\draw[fill=mgreen!10,rounded corners, line width=4pt] (-6,-3.5) rectangle (25, 4.5);
	\draw[fill=red!10,rounded corners, line width=4pt] (-6,-3.5) rectangle (25, -7.15);
		
		
		\newcommand{\networkLayer}[6]{
			\def\a{#1} 
			\def\b{0.02}
			\def\c{#2} 
			\def\t{#3} 
			\def\d{#4} 
			
			\draw[line width=1mm](\c+\t,0,\d) -- (\c+\t,\a,\d) -- (\t,\a,\d);                                                      
			\draw[line width=1mm](\t,0,\a+\d) -- (\c+\t,0,\a+\d) node[midway,below] {#6} -- (\c+\t,\a,\a+\d) -- (\t,\a,\a+\d) -- (\t,0,\a+\d); 
			\draw[line width=1mm](\c+\t,0,\d) -- (\c+\t,0,\a+\d);
			\draw[line width=1mm](\c+\t,\a,\d) -- (\c+\t,\a,\a+\d);
			\draw[line width=1mm](\t,\a,\d) -- (\t,\a,\a+\d);

			\filldraw[#5] (\t+\b,\b,\a+\d) -- (\c+\t-\b,\b,\a+\d) -- (\c+\t-\b,\a-\b,\a+\d) -- (\t+\b,\a-\b,\a+\d) -- (\t+\b,\b,\a+\d); 
			\filldraw[#5] (\t+\b,\a,\a-\b+\d) -- (\c+\t-\b,\a,\a-\b+\d) -- (\c+\t-\b,\a,\b+\d) -- (\t+\b,\a,\b+\d);

			\filldraw[#5] (\c+\t,\b+0.05,\a-\b+\d-0.1) -- (\c+\t,\b+0.05,\b+\d+0.1) -- (\c+\t,\a-\b-0.05,\b+\d+0.1) -- (\c+\t,\a-\b-0.05,\a-\b+\d-0.1); 
		}

		\node at (-1.5cm,1.5cm) (input) {\includegraphics[width=1.5cm]{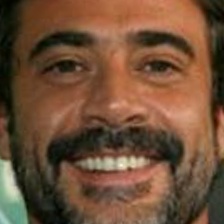}};
		\node at (-3cm,1.5cm)  {\includegraphics[width=1.5cm]{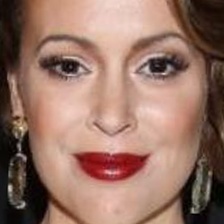}};
		\node at (-4.5cm,1.5cm)  {\includegraphics[width=1.5cm]{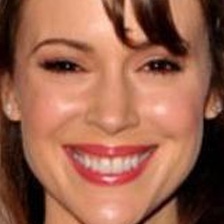}};
		\node at (-1.5cm,3cm)  {\includegraphics[width=1.5cm]{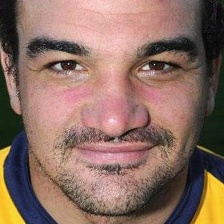}};
		\node at (-3cm,3cm)  {\includegraphics[width=1.5cm]{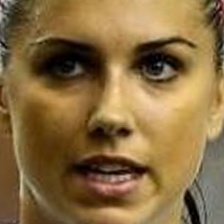}};
		\node at (-4.5cm,3cm)  {\includegraphics[width=1.5cm]{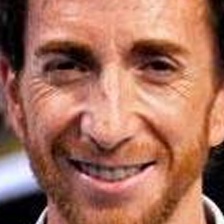}};
		
		\node at (-2,-5.4) (input_test) {\includegraphics[width=3cm]{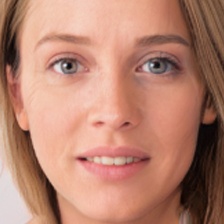}};
		\node[below=0cm of input, xshift=-1.5cm, yshift=0cm, text width=4cm, text centered] {\Large \textbf{Training Data: Real Face Images}};
		\node[right=0cm of input_test, text width=2.5cm] {\Large \textbf{Synthetic Face Image}};

\path [draw,ultra thick,->] (input) -- (0.35,1.5);	
\path [draw,ultra thick,->, dashed] (input_test) |- (0.35,-0.5);

\node[fill=mblue,thick,draw,minimum height=0.35cm,minimum width=1cm,
label=0:Conv + ReLU] at (10,4) {};
\node[fill=morange,thick,draw,minimum height=0.35cm,minimum width=1cm,
label=0:Pooling/Upsampling] at (10,3.5) {};
\node[fill=mgreen,thick,draw,minimum height=0.35cm,minimum width=1cm,
label=0:Activation maps] at (10,3) {};

		\networkLayer{4}{0.2}{2}{0.0}{color=mblue}{}
		\networkLayer{4}{0.2}{2.4}{0.0}{color=morange}{}
		\networkLayer{1.75}{0.5}{4.85}{5.5}{color=mgreen}{112x112x32}
		
		\networkLayer{3}{0.2}{5}{0.0}{color=mblue}{} 
		\networkLayer{3}{0.2}{5.4}{0.0}{color=morange}{} 
		\networkLayer{1}{1}{7.5}{5}{color=mgreen}{56x56x64}
		
		\networkLayer{2}{0.2}{8}{0.0}{color=mblue}{} 
		\networkLayer{2}{0.2}{8.3}{0.0}{color=morange}{}
		\networkLayer{0.5}{1.2}{10.1}{4}{color=mgreen}{28x28x128}
		
		\networkLayer{1}{0.2}{10.6}{0.0}{color=mblue}{} 
		\networkLayer{1}{0.2}{10.8}{0.0}{color=morange}{} 
		\networkLayer{0.5}{0.5}{12.75}{3}{color=mgreen}{28x28x8}

		
		
		\networkLayer{1}{0.2}{13.3}{0.0}{color=mblue}{} 
		\networkLayer{1}{0.2}{13.5}{0.0}{color=morange}{}
		\networkLayer{0.5}{1.2}{15.25}{4}{color=mgreen}{28x28x128}
		
		\networkLayer{2}{0.2}{16}{0.0}{color=mblue}{}       
		\networkLayer{2}{0.2}{16.2}{0.0}{color=morange}{}
		\networkLayer{1}{1}{18.4}{5}{color=mgreen}{56x56x64}
		
		\networkLayer{3}{0.2}{19}{0.0}{color=mblue}{}   
		\networkLayer{3}{0.2}{19.2}{0.0}{color=morange}{} 
		\networkLayer{1.75}{0.5}{21.75}{5.5}{color=mgreen}{112x112x32}

		\networkLayer{4}{0.2}{22}{0.0}{color=mblue}{}   
		\networkLayer{4}{0.2}{22.2}{0.0}{color=morange}{}

		\node at (23.25,-5.4) (output) {\includegraphics[width=3cm]{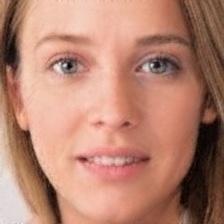}};
		\node[left=0cm of output, text width=5cm] {\Large \textbf{Synthetic Face Image after GANprintR}};		
		\path [draw,ultra thick,->, dashed] (22,0) -| (output);	

\path [draw,ultra thick,->] (2,1)-- (3.75,1);
\path [draw,ultra thick,->] (5,0.75)-- (7.1,0.75);
\path [draw,ultra thick,->] (8,0.5)-- (10.1,0.5);
\path [draw,ultra thick,->] (10.9,0.25)-- (12.75,0.25);
\path [draw,ultra thick,->] (13.65,0.5)-- (15.1,0.5);
\path [draw,ultra thick,->] (16.35,0.75)-- (17.8,0.75);
\path [draw,ultra thick,->] (19,1)-- (20.35,1);

		\node[rotate=90] at (-6.5,0.5) {\Large \textbf{Development Phase}};
		\node[rotate=90] at (-6.5,-5.25) {\Large \textbf{Evaluation Phase}};

	\end{tikzpicture}
	}
	\caption{\textbf{Proposed GAN-fingerprint Removal module (GANprintR) based on a convolutional AutoEncoder (AE).} The AE is trained using only real face images from the development dataset. In the evaluation stage, once the autoencoder is trained, we can pass synthetic face images through it to provide them with additional naturalness, in this way removing the GAN-fingerprint information that may be present in the initial fakes.}
	\label{fig:GAN_fingerprint_Approach}
\end{figure*}
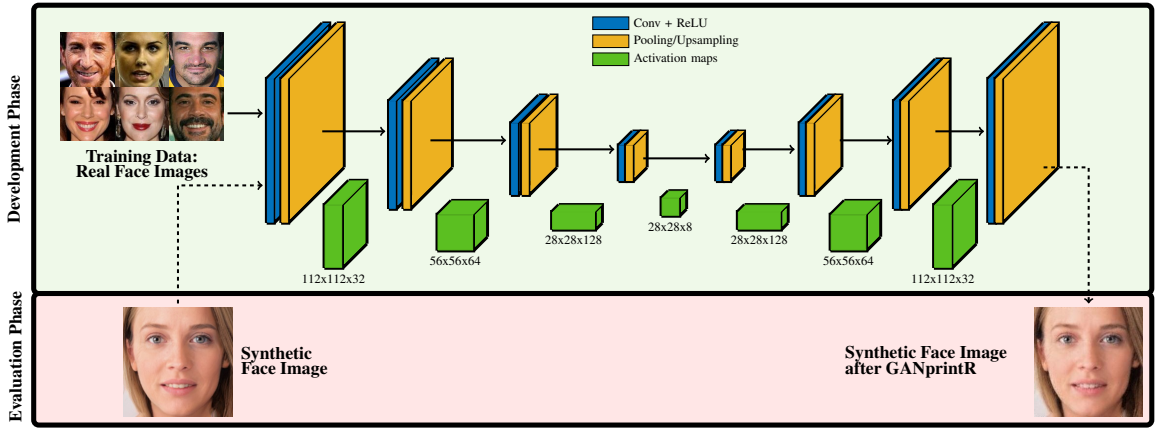

\section{Related Work}\label{related_works}

Various studies have recently evaluated how easy it is to detect manipulations based on the entire face synthesis. Table~\ref{table:relatedWorks} shows a comparison of the most relevant approaches in this area. For each study, we include information related to the features, classifiers, best performance, and databases considered. 

In~\cite{mccloskey2018detecting}, the authors analysed the architecture of GANs in order to detect different artifacts between fake and real images. They proposed a detection system based on colour features and a linear Support Vector Machine (SVM) for the final classification. Their approach achieved a final 70\% Area Under the Curve (AUC) for the best performance when considering the NIST MFC2018 dataset~\cite{NIST_challenge}. A similar approach was followed by Matern \textit{et al.}~\cite{Matern_2019_WACVW} where the authors exploited relatively simple visual artifacts from specific facial regions (e.g., eyes, teeth, facial contours) to detect different types of facial manipulations. In a similar research line, Yang \textit{et al.}~\cite{Yang_2019_ICASSP} exploited the weakness of GANs in generating consistent head poses, and trained a SVM to distinguish between real and synthetic faces based on the estimation of the 3D head pose. 

In~\cite{He_2019_ICIP}, the authors exploited different color channels (YCbCr, HSV and Lab) to extract from a Convolutional Neural Network (CNN) different deep representations, which were subsequently fed to a Random Forest classifier for deciding the realness of an image. 

In~\cite{wang2019fakespotter}, Wang \textit{et al.} conjectured that monitoring neuron behavior could also serve as an asset in detecting fake faces since layer-by-layer neuron activation patterns may capture more subtle features that are important for the facial manipulation detection system. Their proposed approach, named FakeSpoter, extracted as features neuron coverage behaviors of real and fake faces from deep face recognition systems (i.e., VGG-Face~\cite{vggface}, OpenFace~\cite{openface}, and FaceNet~\cite{facenet}), and then trained a SVM for the final classification. The authors tested their proposed approach using real faces from CelebA-HQ~\cite{celebahq} and FFHQ~\cite{Karras_2019_CVPR} databases and synthetic faces created through InterFaceGAN~\cite{shen2019interpreting} and StyleGAN~\cite{Karras_2019_CVPR}, achieving for the best performance a final 84.78\% accuracy for the FaceNet model. 

More recently, Stehouwer \textit{et al.} carried out in~\cite{Jain2019facialManipulation} a complete analysis of different facial manipulation detection methods. They proposed to use attention mechanisms to process and improve the feature maps of CNN models. For the facial manipulation method considered in our study (i.e., entire face synthesis), the authors achieved a final 0.05\% Equal Error Rate (EER) considering real faces from CelebA~\cite{celeba}, FFHQ~\cite{Karras_2019_CVPR}, and FaceForensics++~\cite{rossler2019faceforensics++} databases and fake images created through PGGAN~\cite{pgan} and StyleGAN~\cite{Karras_2019_CVPR} approaches. 

Wang \textit{et al.} carried out in~\cite{wang2019detecting} a very interesting research using publicly available commercial software from Adobe Photoshop in order to synthesise new faces~\cite{adobetool}, and also a professional artist in order to manipulate 50 real photographs. The authors began running a human study through Amazon Mechanical Turk (AMT), showing real and fake images to the participants and asking them to classify each image into one of the classes. The results remark the task difficulty for humans, with a final 53.5\% performance (chance = 50\%). After the human study, the authors proposed an automatic detection system based on Dilated Residual Networks (DRN), achieving Average Precisions (AP) of 99.8\% and 97.4\% for automatic and manual face synthesis manipulation detection. 

In another line of research, some authors have recently focused on the problem of finding the GAN architecture used for generating a specific image potentially synthetic~\cite{Albright_2019_CVPRW,Marra_2019_MIPR}. Yu \textit{et al.} analysed in~\cite{yu2018attributing} the existence and uniqueness of GAN fingerprints to detect fake images. In particular, they proposed a learning-based formulation consisting of an attribution network architecture to map an input image to its corresponding fingerprint image. Therefore, they learned a model fingerprint for each source (each GAN instance plus the real world), such that the correlation index between one image fingerprint and each model fingerprint serves as softmax logit for classification. Their proposed approach was tested using real faces from CelebA database~\cite{celeba} and synthetic faces created through different GAN approaches (PGGAN~\cite{pgan}, SNGAN~\cite{sngan}, CramerGAN~\cite{cramergan}, and MMDGAN~\cite{mmdgan}), achieving a final 99.50\% accuracy for the best performance in manipulation detection.

Finally, we also include for completeness some relevant references to other recent studies focused on the detection of general GAN-based image manipulations, not facial ones:~\cite{Marra_MIPR_2018,zhang2019detecting,huh2018fighting,zhou2018learning,nataraj2019detecting}.


\section{Proposed Approach: \\ GAN-Fingerprint Removal (GANprintR)}\label{proposed_Approach}

Our approach aims at transforming synthetic face images, such that their visual appearance is unaltered but the GAN fingerprints (the discriminative information that permits the distinction from real imagery) are removed. 
Considering that the fingerprints are high frequency signals~\cite{Marra_2019_MIPR}, we hypothesise that their removal could be performed by an autoencoder, which acts as a non-linear low-pass filter. We claim that by using this strategy, the detection capability of state-of-the-art facial manipulation detection methods significantly decreases, while at the same time humans still are not capable of perceiving that images were transformed.

In general, an autoenconder comprises two distinct networks, encoder $\psi$ and decoder $\gamma$:


\begin{align}
\begin{split}\label{eq:1}
&\psi: X \mapsto l \\
&\gamma: l \mapsto X'
\end{split}
\end{align}

\noindent where $X$ denotes the input image to the network, $l$ is the latent feature representation of the input image after passing through the encoder $\psi$, and $X'$ is the reconstructed image generated from $l$, after passing through the decoder $\gamma$. The networks $\psi$ and $\gamma$ can be learned by minimising the reconstruction loss $L_{\psi,\gamma}(X,X') = ||X - X'||^2$ over a development dataset following an iterative learning strategy.

As result, when $L$ is nearly 0, $\psi$ is able to discard all redundant information from $X$ and code it properly into $l$. However, for a reduced size of the latent feature representation vector, $L$ will increase and $\psi$ will be forced to encode in $l$ only the most representative information of $X$. We claim this kind of autoencoder acts as a GAN-fingerprint removal system.

Fig.~\ref{fig:GAN_fingerprint_Approach} describes our proposed approach based on a convolutional AutoEncoder (AE) composed of a sequence of 3$\times$3 convolutional filters, coupled with ReLU activation functions. After each convolutional layer, a 2$\times$2 max-pooling layer is used to progressively decrease the size of the activation map to 28$\times$28$\times$8, which represents the bottleneck of the reconstruction model.

The AE is trained with images from a public dataset that comprises face imagery from real persons. In the evaluation phase, the AE is used to generate improved fakes from input fake faces where GAN ``fingerprints", if present in the initial fakes, will be reduced. The main rationale of this strategy is that by training with real images the AE can learn the core structure of this type of natural data, which can then be exploited to improve existing fakes.

\section{Databases}\label{sec:databases}
Four different public databases and one generated are considered in the experimental framework. Fig.~\ref{datasets_samples} shows some examples of each database. We now summarise the most important features.

\subsection{Real Face Images}\label{subsec:real_databases}

\subsubsection{CASIA-WebFace~\cite{yi2014learning}}\label{subsec:CASIA_WebFace}
this database contains 494,414 face images from 10,575 actors and actresses of IMDb. Face images comprise random pose variations, illumination, facial expression, and resolution.

\subsubsection{VGGFace2~\cite{cao2018vggface2}}\label{subsec:VGGFace2}
this database contains 3.31 million images from 9,131 different subjects, with an average of 363 images per subject. Images were downloaded from the Internet and contain large variations in pose, age, illumination, ethnicity, and profession (e.g., actors, athletes, and politicians).

\subsection{Synthetic Face Images}\label{subsec:fake_databases}

\subsubsection{TPDNE}\label{subsec_TPDNE}
this database comprises 150,000 unique faces, collected from the website\footnote{\url{https://thispersondoesnotexist.com}}. Synthetic images are based on the recent StyleGAN approach~\cite{Karras_2019_CVPR} trained with FFHQ database~\cite{FFHQ}.


\subsubsection{100K-Faces~\cite{100kfaces}}\label{subsec_100K}
this database contains 100,000 synthetic images generated using StyleGAN~\cite{Karras_2019_CVPR}. In this database the StyleGAN network was trained using around 29,000 photos of 69 different models, producing face images with a flat background.
 
\subsubsection{PGGAN~\cite{pgan}}\label{subsec_PGGAN}
this database comprises 80,000 synthetic face images generated using the PGGAN network. In particular, we consider the publicly available model trained using the CelebA-HQ database.

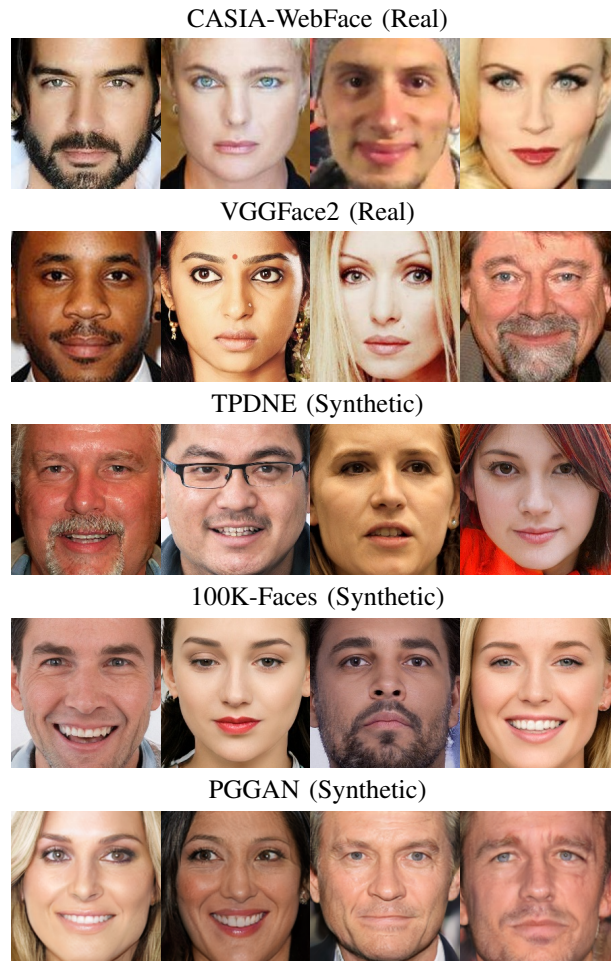
\begin{figure}[t]
\centering
\begin{tikzpicture}[every node/.style={inner sep=0,outer sep=0}]
\node[anchor=north west] at (0,0) (D1_1) {\includegraphics[width=0.225\columnwidth]{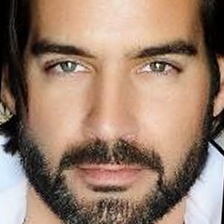}};
\node[right=0cm of D1_1] (D1_2) {\includegraphics[width=0.225\columnwidth]{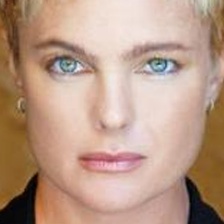}};
\node[right=0cm of D1_2] (D1_3) {\includegraphics[width=0.225\columnwidth]{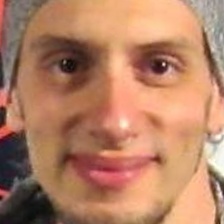}};
\node[right=0cm of D1_3] (D1_4) {\includegraphics[width=0.225\columnwidth]{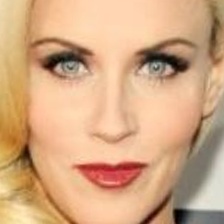}};
\node[above=0.1cm of D1_2,xshift=0.1225\columnwidth] (D1_0) {CASIA-WebFace (Real)};

\node[anchor=north west] at (0,-0.29\columnwidth) (D2_1) {\includegraphics[width=0.225\columnwidth]{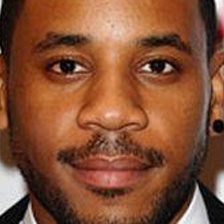}};
\node[right=0cm of D2_1] (D2_2) {\includegraphics[width=0.225\columnwidth]{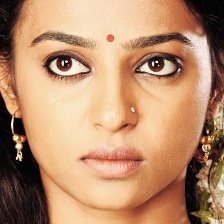}};
\node[right=0cm of D2_2] (D2_3) {\includegraphics[width=0.225\columnwidth]{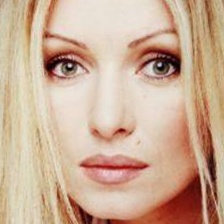}};
\node[right=0cm of D2_3] (D2_4) {\includegraphics[width=0.225\columnwidth]{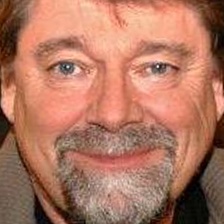}};
\node[above=0.1cm of D2_2,xshift=0.1225\columnwidth] (D2_0) {VGGFace2 (Real)};

\node[anchor=north west] at (0,-0.58\columnwidth) (D3_1) {\includegraphics[width=0.225\columnwidth]{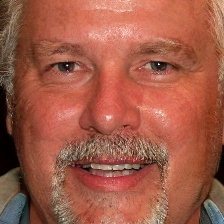}};
\node[right=0cm of D3_1] (D3_2) {\includegraphics[width=0.225\columnwidth]{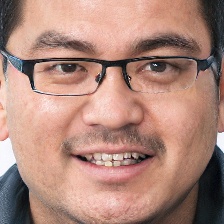}};
\node[right=0cm of D3_2] (D3_3) {\includegraphics[width=0.225\columnwidth]{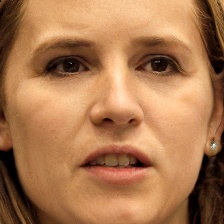}};
\node[right=0cm of D3_3] (D3_4) {\includegraphics[width=0.225\columnwidth]{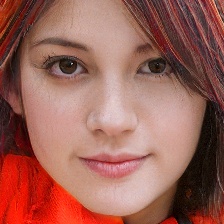}};
\node[above=0.1cm of D3_2,xshift=0.1225\columnwidth] (D3_0) {TPDNE (Synthetic)};

\node[anchor=north west] at (0,-0.87\columnwidth) (D4_1) {\includegraphics[width=0.225\columnwidth]{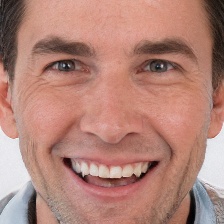}};
\node[right=0cm of D4_1] (D4_2) {\includegraphics[width=0.225\columnwidth]{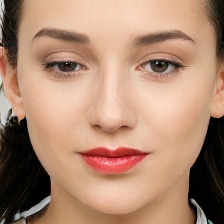}};
\node[right=0cm of D4_2] (D4_3) {\includegraphics[width=0.225\columnwidth]{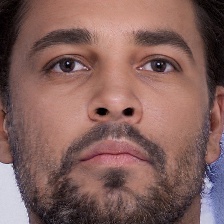}};
\node[right=0cm of D4_3] (D4_4) {\includegraphics[width=0.225\columnwidth]{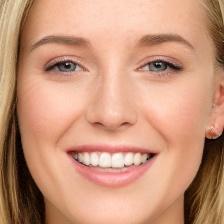}};
\node[above=0.1cm of D4_2,xshift=0.1225\columnwidth] (D4_0) {100K-Faces (Synthetic)};

\node[anchor=north west] at (0,-1.16\columnwidth) (D5_1) {\includegraphics[width=0.225\columnwidth]{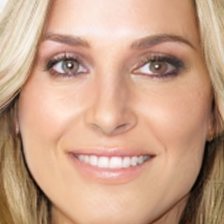}};
\node[right=0cm of D5_1] (D5_2) {\includegraphics[width=0.225\columnwidth]{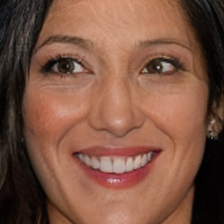}};
\node[right=0cm of D5_2] (D5_3) {\includegraphics[width=0.225\columnwidth]{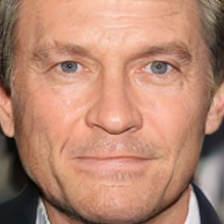}};
\node[right=0cm of D5_3] (D5_4) {\includegraphics[width=0.225\columnwidth]{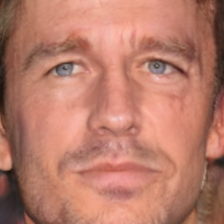}};
\node[above=0.1cm of D5_2,xshift=0.1225\columnwidth] (D5_0) {PGGAN (Synthetic)};

\end{tikzpicture}
\caption{\textbf{Examples of the databases considered in our experiments after applying the pre-processing stage described in Sec.~\ref{sub:preprocessing}}. 
\label{datasets_samples}}
\end{figure}

\section{Experimental Setup}\label{experimental_protocol}

\subsection{Pre-Processing}\label{sub:preprocessing}

In order to ensure fairness in our experimental validation, we created a curated version of all the datasets where the confounding variables were removed. Two different factors were considered in this study:

\begin{itemize}
\item \textit{Background}: this is a clearly distinctive aspect among real and synthetic face images as different acquisition conditions are considered in each database.   
\item \textit{Head pose}: images generated by GANs hardly ever produce high variation from the frontal pose~\cite{Jain2019facialManipulation}, contrasting with most popular real face databases such as CASIA-WebFace and VGGFace2. Therefore, this factor may falsely improve the performance of the detection systems since non-frontal images are more likely to be real faces.
\end{itemize}

To remove these factors from both the real and synthetic images, we extracted 68 face landmarks, using the method described in~\cite{Kazemi_2014_CVPR}. Given the landmarks of the eyes, an affine transformation was determined such that the location of the eyes appears in all images at the same distance from the borders. This step allowed to remove all the background information of the images while keeping the maximum amount of the facial regions. Regarding the head pose, landmarks were used to estimate the pose (\emph{frontal} vs. \emph{non-frontal}). In our experimental framework, we kept only the frontal face images, in order to avoid biased results. After this pre-processing stage, we were able to provide images of constant size (224$\times$224 pixels) as input to the systems. Fig.~\ref{datasets_samples} shows examples of the crop-out faces of each database after applying the pre-processing steps. The synthetic images obtained by this pre-processing stage are the ones used to create the database iFakeFaceDB after being processed by our GANprintR approach.

\subsection{Facial Manipulation Detection Systems}\label{sub:detectionSystems}
Three different state-of-the-art manipulation detection approaches are considered in this study.

1) \textit{XceptionNet}~\cite{chollet2017xception}: this network was selected, essentially because it provides the best detection results in the most recently published studies~\cite{Jain2019facialManipulation,rossler2019faceforensics++,dolhansky2019deepfake}. We followed the same training approach considered in~\cite{rossler2019faceforensics++}: \textit{i)} the model was initialized with the weights obtained after training with the ImageNet dataset~\cite{deng2009imagenet}, \textit{ii)} we changed the last fully-connected layer of the ImageNet model by a new one (two classes, real or synthetic image), \textit{iii)} we fixed all weights up to the final layers and pre-trained the network for few epochs, and finally \textit{iv)} we trained the network for 20 more epochs and chose the best performing model based on validation accuracy. 

2) \textit{Steganalysis}~\cite{nataraj2019detecting}: the method by Nataraj \textit{et al.} was selected for providing an approach based on steganalysis, rather than directly extracting features from the images, as in the XceptionNet approach. In particular, this approach calculates the co-occurrence matrices directly from the image pixels on each channel (red, green and blue), and passes this information through a custom CNN, which allows the network to extract non-linear robust features. Considering that the source code is not available from the authors, we replicated this technique to perform our experiments.

3) \textit{Local Artifacts}~\cite{Matern_2019_WACVW}: we have chosen the method of Matern \textit{et al.}, because it provides an approach based on the direct analysis of the visual facial artifacts, in opposition to the remaining approaches that follow holistic strategies. In particular, the authors of that work claim that some parts of the face (e.g., eyes, teeth, facial contours) provide useful information about the authenticity of the image, and thus train a classifier to distinguish between real and synthetic face images using features extracted from these facial regions.

All our experiments were implemented under a PyTorch framework, with a NVIDIA Titan X GPU. The training of the Xception network was performed using the Adam optimiser with a learning rate of $10^{-3}$, dropout for model regularization with a rate of $0.5$, and a binary cross-entropy loss function. Regarding the steganalysis approach, we reused the parameters adopted for Xception network, since the authors of~\cite{nataraj2019detecting} did not detail the training strategy adopted. 
Regarding the local artifacts approach, we adopted the strategy for detecting ``generated faces", where a k-nearest neighbour classifier is used to distinguish between real and synthetic face images based on eye color features.  

\begin{table*}[t]
\centering
\caption{\textbf{Controlled and in-the-wild scenarios:} manipulation detection performance in terms of EER and Recall (\%) for different development and evaluation setups. $R_{real}$ and $R_{fake}$ denote the Recall of the real and fake classes, respectively. Controlled (Exp. A.1-A.6). In-the-wild (Exp. B.1-B.24). VF2 = VGGFace2. CASIA = CASIA-WebFace. All metrics are given in (\%).}
\scalebox{0.84}{
\begin{tabular}{c|cccc|ccc|ccc|ccc}
& \multicolumn{2}{c}{\textbf{Development}} & \multicolumn{2}{c|}{\textbf{Evaluation}} & \multicolumn{3}{c|}{\textbf{XceptionNet}~\cite{chollet2017xception}} & \multicolumn{3}{c|}{\textbf{Steganalysis}~\cite{nataraj2019detecting}} & \multicolumn{3}{c}{\textbf{Local Artifacts}~\cite{Matern_2019_WACVW}} \\ \hline
\textbf{Experiment} & Real & Synthetic & Real & Synthetic & EER & $R_{real}$ & $R_{fake}$ & EER & $R_{real}$ & $R_{fake}$ & EER & $R_{real}$ & $R_{fake}$\\ \hlineB{3}
\rowcolor{orange!50}
A.1 & VF2 & TPDNE & VF2 & TPDNE & 0.22 & 99.77 & 99.80 &10.92 & 89.07 & 89.10 &38.53 & 60.72 & 62.20 \\ \hlineB{4} 
B.1 & VF2 & TPDNE & VF2 & 100F & 0.45 & 99.30 & 99.80 &23.07 & 71.66 & 85.59 &35.86 & 64.13 & 64.16 \\ \hlineB{4} 
B.2 & VF2 & TPDNE & VF2 & PGGAN & 13.82 & 78.44 & 99.73 &27.12 & 67.28 & 83.87 &40.10 & 59.05 & 60.80 \\ \hlineB{4} 
B.3 & VF2 & TPDNE & CASIA & 100F & 0.35 & 99.30 & 100.00 &24.00 & 71.23 & 83.53 &35.61 & 64.05 & 64.69 \\ \hlineB{4} 
B.4 & VF2 & TPDNE & CASIA & PGGAN & 13.72 & 78.47 & 100.00 &28.05 & 66.81 & 81.61 & 39.87 & 59.0 & 61.4 \\ \hlineB{4} 
\rowcolor{orange!50}
A.2 & VF2 & 100F & VF2 & 100F & 0.28 & 99.70 & 99.73 &12.28 & 87.70 & 87.73 &31.45 & 67.83 & 69.26 \\ \hlineB{4} 
B.5 & VF2 & 100F & VF2 & TPDNE & 21.18 & 70.32 & 99.54 &28.02 & 66.72 & 82.09 &42.89 & 55.17 & 60.16 \\ \hlineB{4} 
B.6 & VF2 & 100F & VF2 & PGGAN & 44.43 & 52.96 & 97.71 &32.62 & 62.35 & 79.31 &48.70 & 50.53 & 52.87 \\ \hlineB{4} 
B.7 & VF2 & 100F & CASIA & TPDNE & 21.07 & 70.37 & 99.94 &28.85 & 66.29 & 80.14 &46.04 & 52.50 & 55.98 \\ \hlineB{4} 
B.8 & VF2 & 100F & CASIA & PGGAN & 44.32 & 53.01 & 99.71 &33.45 & 61.90 & 77.15 & 51.89 & 47.8 & 48.6 \\ \hlineB{4} 
\rowcolor{orange!50}
A.3 & VF2 & PGGAN & VF2 & PGGAN & 0.02 & 99.97 & 100.00 &3.32 & 96.67 & 96.70 &35.13 & 64.33 & 65.41 \\ \hlineB{4} 
B.9 & VF2 & PGGAN & VF2 & TPDNE & 16.85 & 74.79 & 100.00 &33.32 & 60.42 & 91.74 &40.84 & 57.55 & 61.17 \\ \hlineB{4} 
B.10 & VF2 & PGGAN & VF2 & 100F & 5.85 & 89.53 & 100.00 &25.60 & 66.87 & 94.04 &44.47 & 53.99 & 57.77 \\ \hlineB{4} 
B.11 & VF2 & PGGAN & CASIA & TPDNE & 16.85 & 74.79 & 100.00 &35.73 & 59.19 & 81.85 &39.89 & 58.02 & 62.82 \\ \hlineB{4} 
B.12 & VF2 & PGGAN & CASIA & 100F & 5.85 & 89.53 & 100.00 &28.02 & 65.73 & 86.50 & 43.53 & 54.5 & 59.5 \\ \hlineB{4} 
\rowcolor{orange!50}
A.4 & CASIA & TPDNE & CASIA & TPDNE & 0.02 & 99.97 & 100.00 &12.08 & 87.90 & 87.93 &39.36 & 59.62 & 61.65 \\ \hlineB{4} 
B.13 & CASIA & TPDNE & VF2 & 100F & 1.75 & 99.35 & 97.20 &36.68 & 59.58 & 71.82 &39.03 & 60.67 & 61.25 \\ \hlineB{4} 
B.14 & CASIA & TPDNE & VF2 & PGGAN & 4.42 & 94.21 & 97.04 &30.77 & 65.13 & 76.40 &38.94 & 61.02 & 61.10 \\ \hlineB{4} 
B.15 & CASIA & TPDNE & CASIA & 100F & 0.32 & 99.37 & 100.00 &34.12 & 61.02 & 78.41 &38.05 & 61.20 & 62.67 \\ \hlineB{4} 
B.16 & CASIA & TPDNE & CASIA & PGGAN & 2.98 & 94.37 & 100.00 &28.20 & 66.48 & 82.19 & 37.96 & 61.5 & 62.5 \\ \hlineB{4} 
\rowcolor{orange!50}
A.5 & CASIA & 100F & CASIA & 100F & 0.08 & 99.90 & 99.93 &16.05 & 83.94 & 83.96 &33.96 & 65.04 & 67.03 \\ \hlineB{4} 
B.17 & CASIA & 100F & VF2 & TPDNE & 5.93 & 97.69 & 90.95 &34.00 & 62.64 & 71.80 &43.11 & 55.00 & 59.83 \\ \hlineB{4} 
B.18 & CASIA & 100F & VF2 & PGGAN & 10.08 & 89.64 & 90.20 &45.63 & 52.91 & 58.71 &46.36 & 52.37 & 55.92 \\ \hlineB{4} 
B.19 & CASIA & 100F & CASIA & TPDNE & 1.10 & 97.91 & 99.93 &31.67 & 63.97 & 76.67 &44.22 & 53.94 & 58.54 \\ \hlineB{4} 
B.20 & CASIA & 100F & CASIA & PGGAN & 5.25 & 90.55 & 99.93 &43.30 & 54.34 & 64.74 & 47.49 & 51.3 & 54.6 \\ \hlineB{4} 
\rowcolor{orange!50}
A.6 & CASIA & PGGAN & CASIA & PGGAN & 0.05 & 99.93 & 99.97 &4.62 & 95.37 & 95.40 &34.79 & 64.42 & 66.00 \\ \hlineB{4} 
B.21 & CASIA & PGGAN & VF2 & TPDNE & 4.90 & 99.96 & 91.10 &31.73 & 61.93 & 88.92 &43.52 & 55.25 & 57.94 \\ \hlineB{4} 
B.22 & CASIA & PGGAN & VF2 & 100F & 4.88 & 100.00 & 91.10 &41.97 & 54.63 & 80.35 &44.69 & 54.05 & 56.89 \\ \hlineB{4} 
B.23 & CASIA & PGGAN & CASIA & TPDNE & 0.03 & 99.97 & 99.97 &31.43 & 62.08 & 90.07 &41.46 & 56.64 & 61.00 \\ \hlineB{4} 
B.24 & CASIA & PGGAN & CASIA & 100F & 0.02 & 100.00 & 99.97 &41.67 & 54.79 & 82.22 & 42.63 & 55.5 & 60.0 \\ \hlineB{4} 

\end{tabular}
}
\label{table:intra_inter_experiments}
\end{table*}

\subsection{Protocol}\label{sub:experimentalProtocol}

The experimental protocol designed in this study aims at performing an exhaustive analysis of state-of-the-art facial manipulation detection systems. As such, three different experiments are considered: \textit{i)} controlled scenarios, \textit{ii)} in-the-wild scenarios, and \textit{iii)} GAN-fingerprint removal.

Each database was divided into two disjoint datasets, one for the development of the systems (70\%) and the other one for evaluation purposes (30\%). Additionally, the development dataset was divided into two disjoint subsets, training (75\%) and validation (25\%). The same number of real and synthetic images were considered in the experimental framework. In addition, for real face images, different users were considered in the development and evaluation datasets, in order to avoid biased results.

Our proposed GANprintR was trained during 100 epochs, using the Adam optimizer with a learning rate of $10^{-3}$, and mean square error (MSE) to obtain the reconstruction loss. To ensure an unbiased evaluation, our GANprintR was trained with images from the MS-Celeb dataset~\cite{guo2016msceleb}, since it is disjoint from the datasets used in the development and evaluation of all the fake detection systems used in our experiments.

\begin{table*}
\centering
\caption{\textbf{Comparison between the proposed approach (GANprintR) and typical image manipulations.} The detection performance is provided in terms of EER and Recall (\%) for experiments A.1 to A.6, when using different versions of the evaluation set. TED stands for transformation of the evaluation data and details the technique used to modify the test set before fake detection. $R_{real}$ and $R_{fake}$ denote the Recall of the real and fake classes, respectively. }
\scalebox{0.85}{
\begin{tabular}{c|cccc|lccccc}
& \multicolumn{2}{c}{\textbf{Development}} & \multicolumn{2}{c|}{\textbf{Evaluation}} & & \multicolumn{3}{c}{\textbf{XceptionNet}~\cite{chollet2017xception}} \\ \hline
\textbf{Experiment} & Real & Synthetic & Real & Synthetic & TED & EER(\%) & $R_{real}$(\%) & $R_{fake}$(\%) & PSNR(db) & SSIM \\ \hlineB{5} 
\rowcolor{orange!50}
\multirow{1}{*}{A.1} & \multirow{ 1}{*}{VF2} & \multirow{ 1}{*}{TPDNE} & \multirow{ 1}{*}{VF2} & \multirow{ 1}{*}{TPDNE} & Original & 0.22 & 99.77 & 99.80 & - & - \\ \hlineB{5} 
 & & & & & Downsize & 1.17 & 98.83 & 98.87 & 35.55 & 0.93 \\ 
 & & & & & Low-Pass Filter & 0.83 & 99.17 & 99.20 & 34.63 & 0.92 \\ 
 & & & & & JPEG Compression & 1.53 & 98.47 & 98.50 & 36.02 & 0.96 \\ 
 & & & & & GANprintR &10.63 & 89.37 & 89.40 & 35.01 & 0.96 \\ \hlineB{5}  
\rowcolor{orange!50}
\multirow{1}{*}{A.2} & \multirow{ 1}{*}{VF2} & \multirow{ 1}{*}{100F} & \multirow{ 1}{*}{VF2} & \multirow{ 1}{*}{100F} & Original & 0.28 & 99.70 & 99.73 & - & - \\ \hlineB{5} 
 & & & & & Downsize & 0.87 & 99.13 & 99.17 & 36.24 & 0.95 \\ 
 & & & & & Low-Pass Filter & 2.87 & 97.10 & 97.13 & 35.22 & 0.93 \\ 
 & & & & & JPEG Compression & 1.83 & 98.17 & 98.20 & 36.76 & 0.97 \\ 
 & & & & & GANprintR &6.37 & 93.64 & 93.66 & 35.59 & 0.96 \\ \hlineB{5}  
\rowcolor{orange!50}
\multirow{1}{*}{A.3} & \multirow{ 1}{*}{VF2} & \multirow{ 1}{*}{PGGAN} & \multirow{ 1}{*}{VF2} & \multirow{ 1}{*}{PGGAN} & Original & 0.02 & 99.97 & 100.00 & - & - \\ \hlineB{5} 
 & & & & & Downsize & 3.70 & 96.27 & 96.30 & 34.85 & 0.91 \\ 
 & & & & & Low-Pass Filter & 1.53 & 98.43 & 98.47 & 34.10 & 0.90 \\ 
 & & & & & JPEG Compression & 30.93 & 69.04 & 69.06 & 35.85 & 0.96 \\ 
 & & & & & GANprintR &17.27 & 82.71 & 82.73 & 34.82 & 0.95 \\ \hlineB{5}  
\rowcolor{orange!50}
\multirow{1}{*}{A.4} & \multirow{ 1}{*}{CASIA} & \multirow{ 1}{*}{TPDNE} & \multirow{ 1}{*}{CASIA} & \multirow{ 1}{*}{TPDNE} & Original & 0.02 & 99.97 & 100.00 & - & - \\ \hlineB{5} 
 & & & & & Downsize & 1.00 & 98.97 & 99.00 & 35.55 & 0.93 \\ 
 & & & & & Low-Pass Filter & 0.07 & 99.90 & 99.93 & 34.63 & 0.92 \\ 
 & & & & & JPEG Compression & 2.50 & 97.47 & 97.50 & 36.02 & 0.96 \\ 
 & & & & & GANprintR &4.47 & 95.50 & 95.53 & 35.01 & 0.96 \\ \hlineB{5}  
\rowcolor{orange!50}
\multirow{1}{*}{A.5} & \multirow{ 1}{*}{CASIA} & \multirow{ 1}{*}{100F} & \multirow{ 1}{*}{CASIA} & \multirow{ 1}{*}{100F} & Original & 0.08 & 99.90 & 99.93 & - & - \\ \hlineB{5} 
 & & & & & Downsize & 6.27 & 93.70 & 93.73 & 36.24 & 0.95 \\ 
 & & & & & Low-Pass Filter & 11.53 & 88.44 & 88.46 & 35.22 & 0.93 \\ 
 & & & & & JPEG Compression & 3.27 & 96.73 & 96.77 & 36.76 & 0.97 \\ 
 & & & & & GANprintR & 11.47 & 88.50 & 88.53 & 35.59 & 0.96 \\ \hlineB{5}  
\rowcolor{orange!50}
\multirow{1}{*}{A.6} & \multirow{ 1}{*}{CASIA} & \multirow{ 1}{*}{PGGAN} & \multirow{ 1}{*}{CASIA} & \multirow{ 1}{*}{PGGAN} & Original & 0.05 & 99.93 & 99.97 & - & - \\ \hlineB{5} 
 & & & & & Downsize & 7.77 & 92.24 & 92.26 & 34.85 & 0.91 \\ 
 & & & & & Low-Pass Filter & 2.10 & 97.90 & 97.93 & 34.10 & 0.90 \\ 
 & & & & & JPEG Compression & 5.37 & 94.64 & 94.66 & 35.85 & 0.96 \\ 
 & & & & & GANprintR & 8.37 & 91.64 & 91.66 & 34.82 & 0.95 \\  \hline
\end{tabular}
}
\label{table:fingerprint_experiments}
\end{table*}

\section{Experimental Results}\label{experimental_results}

\subsection{Controlled Scenarios}\label{subsec:IntraDatabase}

In this section, we report the results of the detection of entire face synthesis in controlled scenarios, i.e., when samples from the same databases were considered for both development and final evaluation of the detection systems. This is the strategy commonly used in most studies, typically resulting in very good performance (see Sec.~\ref{related_works}).

\begin{figure}[tb]
\centering
\subfigure[XceptionNet~\cite{chollet2017xception}]{\label{fig:loss_TPDNE}
\includegraphics[width=0.48\linewidth]{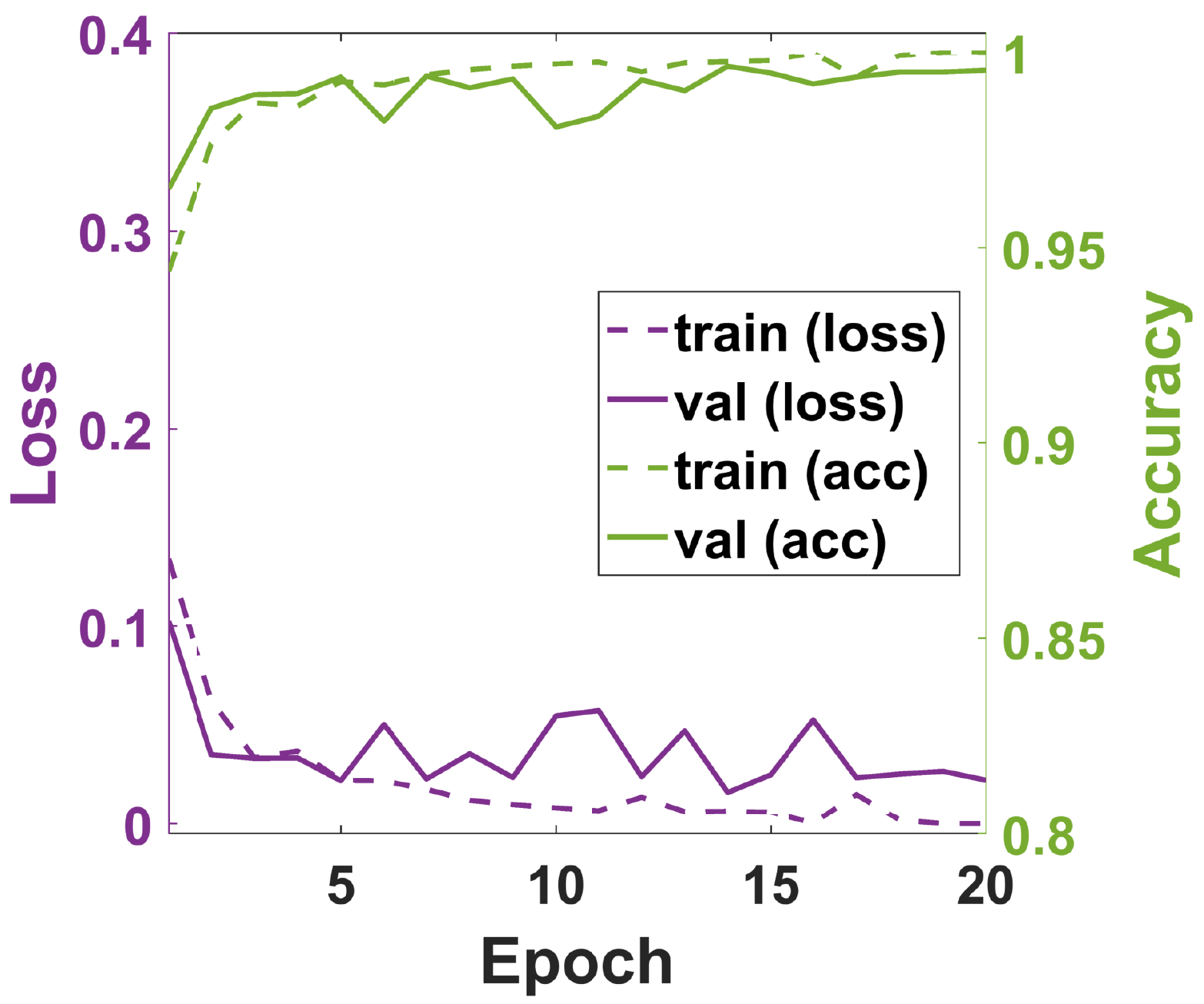}}
\subfigure[Steganalysis~\cite{nataraj2019detecting}]{\label{fig:loss_100KFaces}
\includegraphics[width=0.48\linewidth]{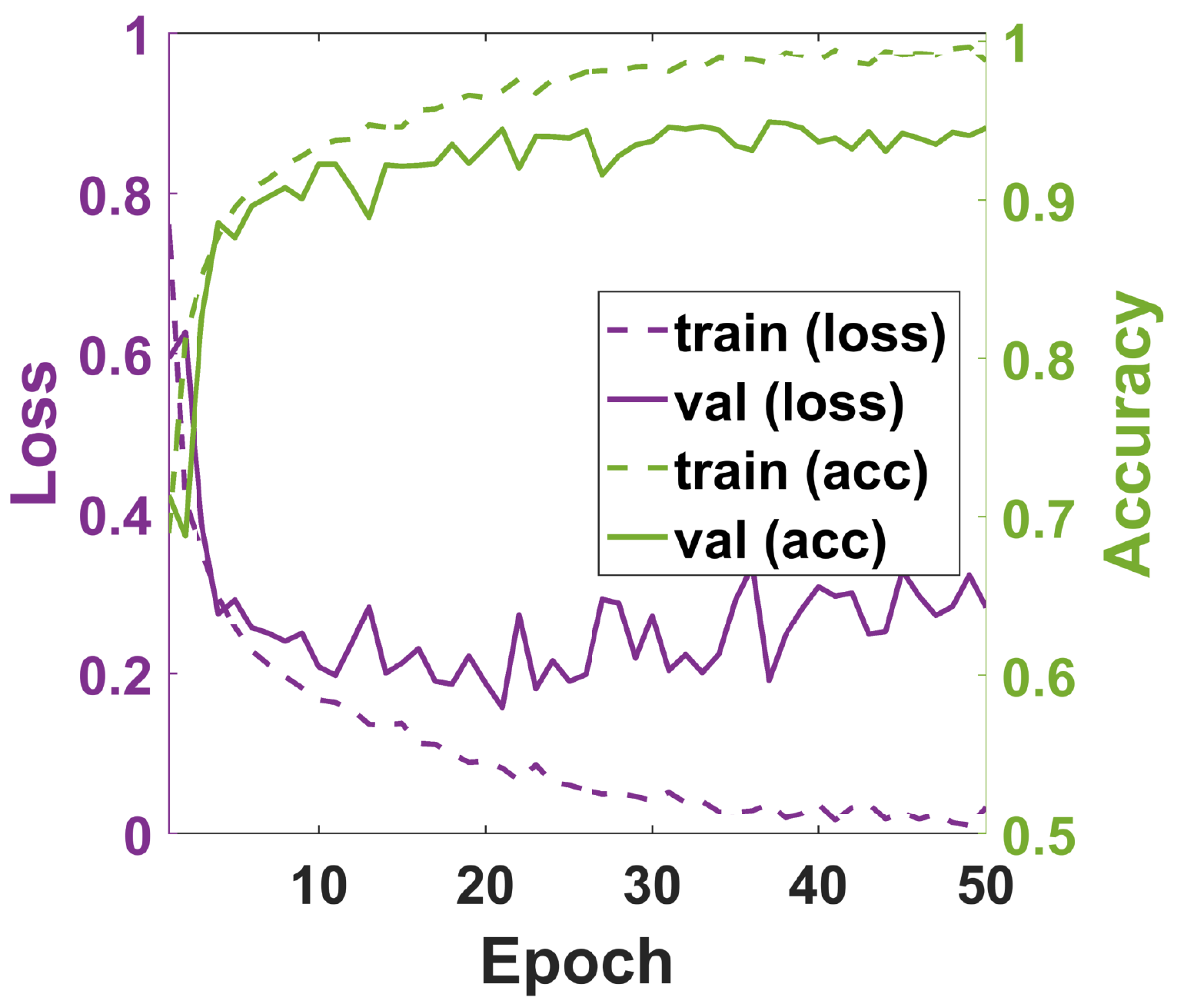}}
\caption{\textbf{Exp. A.1:} evolution of the loss/accuracy with the number of epochs.} 
\label{fig:evolution_intra-analysis}
\end{figure}

A total of six experiments are carried out: A.1 to A.6. Table~\ref{table:intra_inter_experiments} describes the development and evaluation databases considered in each experiment together with the corresponding final evaluation results in terms of EER. Additionally, we represent in Fig.~\ref{fig:evolution_intra-analysis} the evolution of the loss/accuracy of the XceptionNet and Steganalysis detection systems for Exp. A.1.

The analysis of Fig.~\ref{fig:evolution_intra-analysis} shows that both XceptionNet and Steganalysis approaches are able to learn discriminative features to detect between real and synthetic face images. The training process was faster for the XceptionNet detection system compared with Steganalysis, converging to a lower loss value in fewer epochs (close to zero after 20 epochs). The best validation accuracies achieved in Exp. A.1 for the XceptionNet and Steganalysis approaches are 99\% and 95\%, respectively. Similar trends are observed for the other experiments.

We now analyse the results included in Table~\ref{table:intra_inter_experiments} for experiments A.1 to A.6. Analysing the results obtained by the XceptionNet system, almost ideal performance is achieved with EER values less than 0.5\%. These results are in agreement to previous studies in the topic (see Sec.~\ref{related_works}), pointing for the potential of the XceptionNet model in controlled scenarios. Regarding the Steganalysis approach, a higher degradation of the system performance is observed, when compared with the XceptionNet approach, especially for the 100K-Face database, e.g., a 16\% EER is obtained in Exp. A.5. Finally, it can be observed that the approach based on local artifacts was the least efficient to spot the differences between real and synthetic data, with an average 35.5\% EER over all experiments.

In summary, for controlled scenarios XceptionNet has excellent manipulation detection accuracies, then Steganalysis provides good accuracies, and finally Local Artifacts has poor accuracy. In the next section we will see the limitations of these techniques in-the-wild.

\subsection{In-the-Wild Scenarios}\label{subsec:InterDatabase}

This section evaluates the performance of the facial manipulation detection systems in more realistic scenarios, i.e., in-the-wild. The following aspects are considered: \textit{i)} different development and evaluation databases, and \textit{ii)} different image resolution/blur among the development and evaluation of the models. This last point is particularly important, as the quality of raw images/videos is usually modified when, e.g., they are uploaded to social media. The effect of image resolution has been preliminary analysed in previous studies~\cite{rossler2019faceforensics++,korshunov2018deepfakes}, but for different facial manipulation groups, i.e., face swapping/identity swap and facial expression manipulation. The main goal of this section is to analyse the generalisation capability of state-of-the-art entire face synthesis detection in unconstrained scenarios. 

First, we focus on the scenario of considering the same real but different synthetic databases in development and evaluation (Exp. B.1, B.2, B.5, B.6, and so on, provided in Table~\ref{table:intra_inter_experiments}). In general, the results achieved in the experiments evidence a high degradation of the detection performance regardless of the facial manipulation detection approach. For the XceptionNet, the average EER is 11.2\%, i.e., over 20 times higher than the results achieved in Exp. A.1-A.6 ($<$0.5\% average EER). Regarding the Steganalysis approach, the average EER is 32.5\%, i.e., more than 3 times higher than the results achieved in Exp. A.1-A.6 (9.8\% average EER). For Local Artifacts, the observed average EER is 42.4\%, with an average worsening of 19\%. The large degradation of the first two detectors suggests that they might rely heavily on the GAN fingerprints of the training data. This result confirms the hypothesis that different GAN models produce different fingerprints, as also mentioned in previous studies~\cite{yu2018attributing}. Moreover, these results suggest that these GAN fingerprints are the information used by the detectors to distinguish between real and synthetic data.

Table~\ref{table:intra_inter_experiments} also considers the case of using different real and synthetic databases for both development and evaluation (Exp. B.3, B.4, B.7, B.8, etc.). In this scenario, average EERs of 9.3\%, 32.3\% and 42.3\% in fake detection are obtained for XceptionNet, Steganalysis, and Local Artifacts, respectively. When comparing these results with the EERs of the previous experiments (where only the synthetic evaluation set was changed), no significant gap in performance is found, which suggests that the change of synthetic data in training might be the main cause for performance degradation.

Finally, we also analyse how different image transformations affect facial manipulation detection systems. In this analysis, we focus only on the XceptionNet model as it provides much better results when compared with the remaining detection systems. For each baseline experiment (A.1 to A.6), the evaluation set (both real and fake images) was transformed by: \emph{i)} resolution downsizing (1/3 of the original resolution), \emph{ii)} a low-pass filter ($9\times9$ Gaussian kernel, $\sigma=1.7$), and \emph{iii)} jpeg image compression using a quality level of 60. The resulting EER together with the Recall, PSRN, and SSIM values are provided in Table~\ref{table:fingerprint_experiments}, together with the performance of the original images. The results suggest a high performance degradation in manipulation detection for all experiments, proving the vulnerability of fake detection systems to unseen conditions, even if they result from simple image transformations. These findings agree with the conclusions extracted in other studies of the literature. For example, Marra \textit{et al.} evaluated in~\cite{Marra_MIPR_2018} the robustness of different fake detectors over different training and testing scenarios, considering fake images created using image-to-image translations~\cite{zhu2017unpaired}. For the XceptionNet approach and the image compression scenario, the authors achieved an accuracy in manipulation detection of 87.17\%, an average absolute worsening of 7.32\% compared with the uncompressed scenario (accuracy of 94.49\%).

\begin{figure}
\centering
\begin{tikzpicture}[every node/.style={inner sep=0,outer sep=0}]
\node[anchor=north west] (A) at (0,0) {\includegraphics[width=\columnwidth]{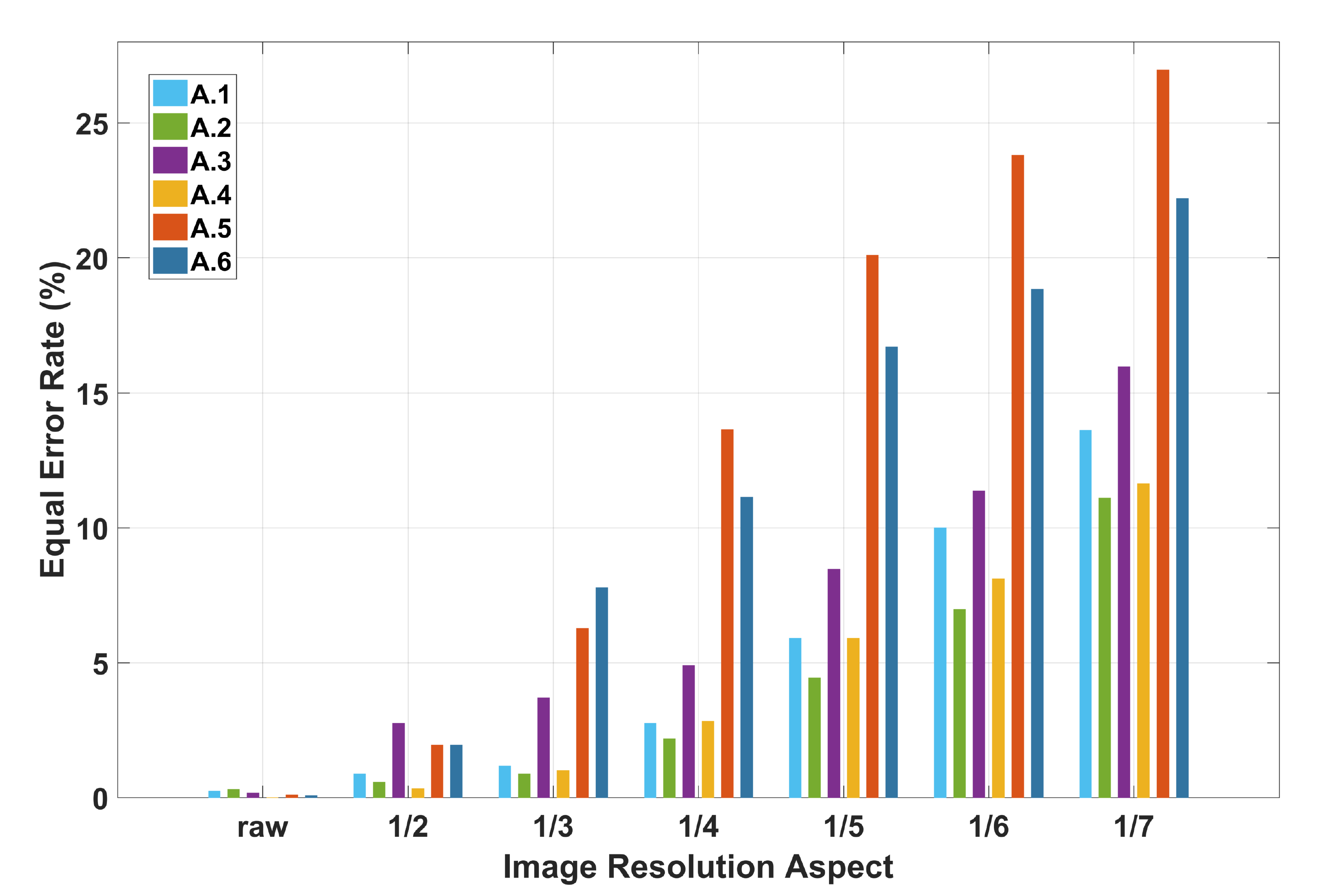}};

\node[below=0cm of A,xshift=-0.45\columnwidth] (I1) {\includegraphics[width=0.14\columnwidth]{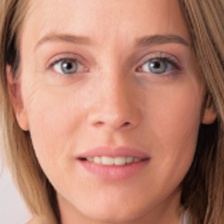}};
\node[below=0.2em of I1] {\tiny raw};

\node[right=0cm of I1] (I2) {\includegraphics[width=0.14\columnwidth]{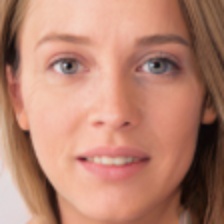}};
\node[below=0.2em of I2] {\tiny 1/2};

\node[right=0cm of I2] (I3) {\includegraphics[width=0.14\columnwidth]{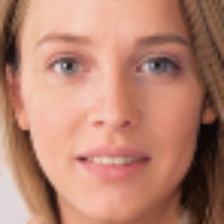}};
\node[below=0.2em of I3] {\tiny 1/3};

\node[right=0cm of I3] (I4) {\includegraphics[width=0.14\columnwidth]{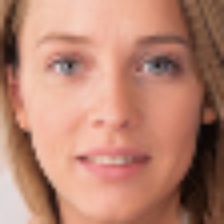}};
\node[below=0.2em of I4] {\tiny 1/4};

\node[right=0cm of I4] (I5) {\includegraphics[width=0.14\columnwidth]{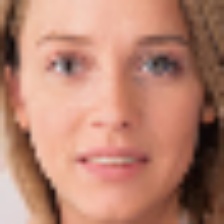}};
\node[below=0.2em of I5] {\tiny 1/5};

\node[right=0cm of I5] (I6) {\includegraphics[width=0.14\columnwidth]{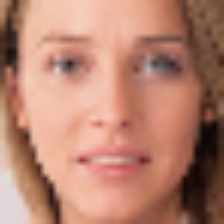}};
\node[below=0.2em of I6] {\tiny 1/6};

\node[right=0cm of I6] (I7) {\includegraphics[width=0.14\columnwidth]{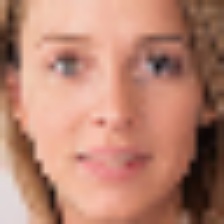}};
\node[below=0.2em of I7] {\tiny 1/7};

\end{tikzpicture}
\caption{\textbf{Robustness of the fake detection system regarding the image resolution.} The XceptionNet model is trained with the raw image resolution and evaluated with lower image resolutions. Note how the EER increases significantly while reducing the image resolution.}
\label{fig:imageSize}
\end{figure}
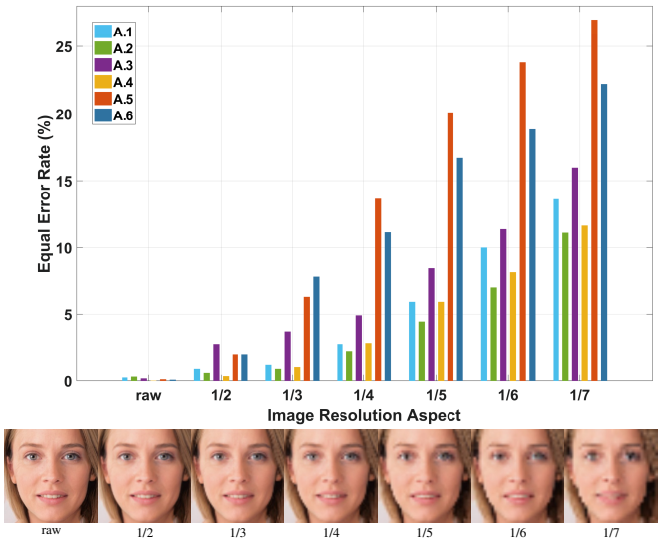

To further understand the impact of these transformations, we evaluated an increasing downsize ratio in the performance of the fake detection system. Fig.~\ref{fig:imageSize} depicts the detection performance results in terms of EER(\%), from lower to higher modifications of the image resolution. In general, we can observe increasingly higher degradation of the fake detection performance for decreasing resolution. For example, when the image resolution is reduced to 1/4, the average EER in manipulation detection increases 6\% when compared with the raw image resolution (raw equals to 1/1). This performance degradation is even higher when we further reduce the image resolution, with EERs(\%) higher than 15\%. These results support the conclusion about a poor generalisation capacity of state-of-the-art facial manipulation detection systems to unseen conditions.


%
%

\subsection{GAN-Fingerprint Removal}\label{subsec:UncontrainedScenrio}

This section analyses the results of the proposed strategy for GAN-fingerprint Removal (GANprintR). We evaluated to what extent our method is capable of spoofing state-of-the-art fake detectors by improving fake images already obtained with some of the best and most realistic known methods for entire face synthesis. For this, the experiments A.1 to A.6 were repeated for the XceptionNet detection system, but the fake images of the evaluation set were transformed after passing through our proposed GANprintR. Table~\ref{table:fingerprint_experiments} provides the results achieved for both the original fake data and after GANprintR. The analysis shows that GANprintR results in higher fake detection error than the remaining attacks, while maintaining a similar or even better visual quality. In all the experiments, the EER of the manipulation detection increases when using GANprintR to transform the synthetic face images. Also, the detection degradation is higher than other types of attacks for similar PSNR values and slightly higher values of SSIM. In particular, the average EER when considering GANprintR is 9.8\%, i.e., over 20 times higher than the results achieved when using the original fakes ($<$0.5\% average EER). This suggests that our method is not simply removing high-frequency information (evidenced by the comparison with the low-pass filter and downsize) but it is also removing the GAN fingerprints from the fakes improving their naturalness. It is important to remark that different real face databases were considered for training the face manipulation detection systems and our GANprintR module.

\begin{figure}
\centering
\begin{tikzpicture}[every node/.style={inner sep=0,outer sep=0}]
\node[anchor=north west] (A) at (0,0) {\includegraphics[width=\columnwidth]{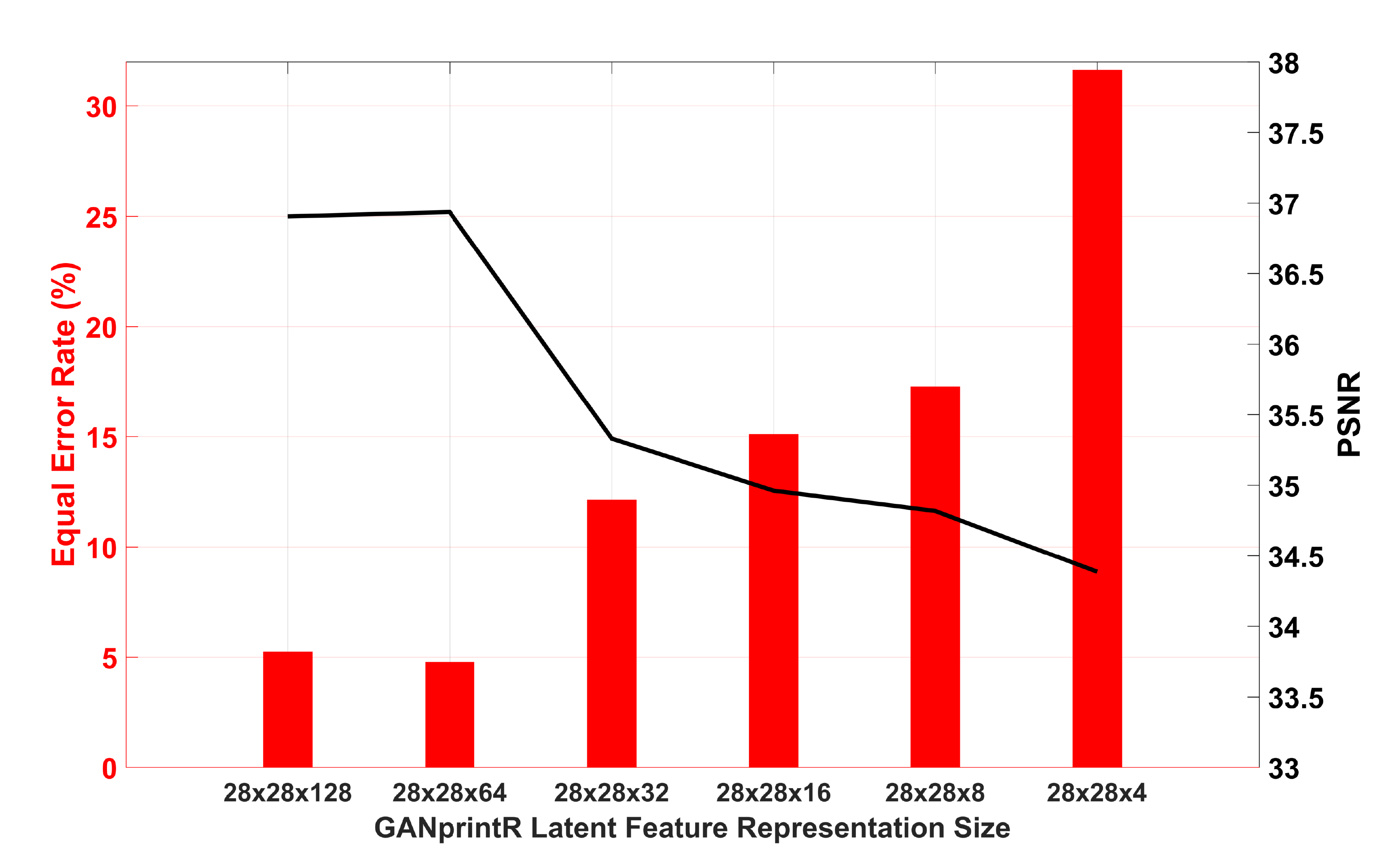}};

\node[below=0cm of A, xshift=-0.45\columnwidth] (I1) {\includegraphics[width=0.14\columnwidth]{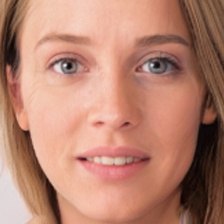}};
\node[below=0.2em of I1] {\tiny original};

\node[right=0cm of I1] (I2) {\includegraphics[width=0.14\columnwidth]{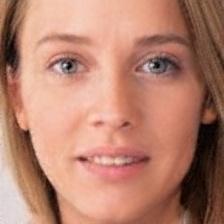}};
\node[below=0.2em of I2] {\tiny 28x28x128};

\node[right=0cm of I2] (I3) {\includegraphics[width=0.14\columnwidth]{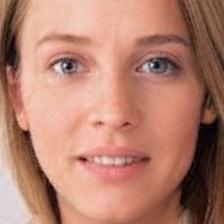}};
\node[below=0.2em of I3] {\tiny 28x28x64};

\node[right=0cm of I3] (I4) {\includegraphics[width=0.14\columnwidth]{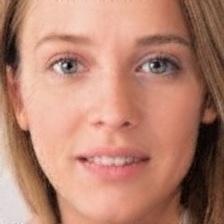}};
\node[below=0.2em of I4] {\tiny 28x28x32};

\node[right=0cm of I4] (I5) {\includegraphics[width=0.14\columnwidth]{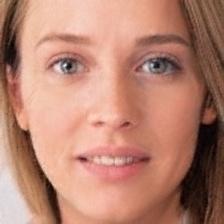}};
\node[below=0.2em of I5] {\tiny 28x28x16};

\node[right=0cm of I5] (I6) {\includegraphics[width=0.14\columnwidth]{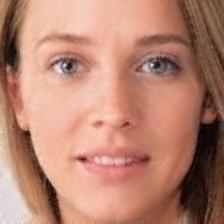}};
\node[below=0.2em of I6] {\tiny 28x28x8};

\node[right=0cm of I6] (I7) {\includegraphics[width=0.14\columnwidth]{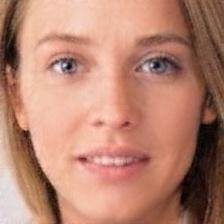}};
\node[below=0.2em of I7] {\tiny 28x28x4};

\end{tikzpicture}
\caption{\textbf{Robustness of the fake detection system after our proposed GAN-fingerprint Removal (GANprintR).} The latent feature representation size of the AE is varied to analyse the impact on both system performance and visual aspect of the reconstructed images. Note how the EER increases significantly when considering our proposed spoof approach, while maintaining a high visual similarity with the original image.}
\label{fig4}
\end{figure}
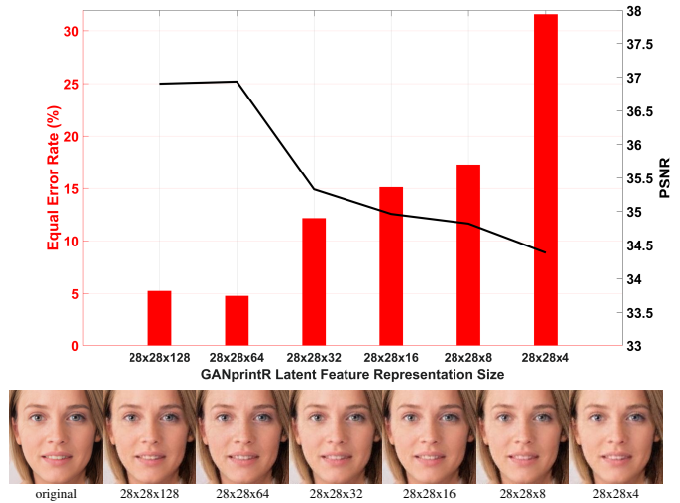

In addition, we provide in Fig.~\ref{fig4} an analysis of the impact of the latent feature representation of the autoencoder in terms of EER and PSNR. In particular, we follow the experimental protocol considered in Exp. A.3, and calculate the EER of XceptionNet for detecting fakes improved with various configurations of GANprintR. Moreover, the PSNR for each set of transformed images is also included in Fig.~\ref{fig4} together with a face example of each configuration to visualise the image quality. The face examples included in Fig.~\ref{fig4} show no substantial differences between the original fake and the resulting fakes after GANprintR for the different latent feature representation size of the GANprintR, which is confirmed by the tight range of PSNR values obtained along the different latent feature representations. On the other hand, EER values of fake detection significantly increase as the size of latent feature representations diminish, evidencing that GANprintR is capable of spoofing state-of-the-art manipulation detection systems without significantly degrading the visual aspect of the image.

\begin{table*}
\caption{\textbf{Impact of the GANprintR approach on three state-of-the-art manipulation detection approaches.} A significant performance degradation is observed in all manipulation detection approaches when exposed to images transformed by the proposed GANprintR approach. The detection performance is provided in terms of EER and Recall (\%), while $R_{real}$ and $R_{fake}$ denote the Recall of the real and fake classes, respectively.\label{tab:ganprintr_alldetectors}}
\scalebox{0.65}{
\begin{tabular}{c|cccc|lccc|ccc|ccc}
& \multicolumn{2}{c}{\textbf{Development}} & \multicolumn{2}{c|}{\textbf{Evaluation}} &  & \multicolumn{3}{c|}{\textbf{XceptionNet}~\cite{chollet2017xception}} & \multicolumn{3}{c|}{\textbf{Steganalysis}~\cite{nataraj2019detecting}} & \multicolumn{3}{c}{\textbf{Local Artifacts}~\cite{Matern_2019_WACVW}} \\ \hline
\textbf{Experiment} & Real & Synthetic & Real & Synthetic & data & EER(\%) & $R_{real}$(\%) & $R_{fake}$(\%) & EER(\%) & $R_{real}$(\%) & $R_{fake}$(\%) & EER(\%) & $R_{real}$(\%) & $R_{fake}$(\%) \\ \hlineB{5} 
\multirow{2}{*}{A.1} & \multirow{2}{*}{VF2} & \multirow{2}{*}{TPDNE} & \multirow{2}{*}{VF2} & \multirow{2}{*}{TPDNE} & Original & 0.22 & 99.77 & 99.80 &10.92 & 89.07 & 89.10 &38.53 & 60.72 & 62.20  \\ 
 & & & & & GANprintR &10.63 & 89.37 & 89.40 & 22.37 & 77.61 & 77.63 & 44.06 & 55.16 & 56.67 \\ \hline 
\multirow{2}{*}{A.2} & \multirow{2}{*}{VF2} & \multirow{2}{*}{100F} & \multirow{2}{*}{VF2} & \multirow{2}{*}{100F} & Original & 0.28 & 99.70 & 99.73 &12.28 & 87.70 & 87.73 &31.45 & 67.83 & 69.26  \\ 
 & & & & & GANprintR &6.37 & 93.64 & 93.66 & 17.30 & 82.71 & 82.73 & 36.35 & 62.93 & 64.41 \\ \hline 
\multirow{2}{*}{A.3} & \multirow{2}{*}{VF2} & \multirow{2}{*}{PGGAN} & \multirow{2}{*}{VF2} & \multirow{2}{*}{PGGAN} & Original & 0.02 & 99.97 & 100.00 &3.32 & 96.67 & 96.70 &35.13 & 64.33 & 65.41  \\ 
 & & & & & GANprintR &17.27 & 82.71 & 82.73 & 35.13 & 64.85 & 64.85 & 42.24 & 57.28 & 58.29 \\ \hline 
\multirow{2}{*}{A.4} & \multirow{2}{*}{CASIA} & \multirow{2}{*}{TPDNE} & \multirow{2}{*}{CASIA} & \multirow{2}{*}{TPDNE} & Original & 0.02 & 99.97 & 100.00 &12.08 & 87.90 & 87.93 &39.36 & 59.62 & 61.65  \\ 
 & & & & & GANprintR &4.47 & 95.50 & 95.53 & 24.97 & 75.04 & 75.06 & 42.75 & 56.16 & 58.37 \\ \hline 
\multirow{2}{*}{A.5} & \multirow{2}{*}{CASIA} & \multirow{2}{*}{100F} & \multirow{2}{*}{CASIA} & \multirow{2}{*}{100F} & Original & 0.08 & 99.90 & 99.93 &16.05 & 83.94 & 83.96 &33.96 & 65.04 & 67.03  \\ 
 & & & & & GANprintR &11.47 & 98.50 & 98.53 & 19.80 & 80.17 & 80.19 & 38.14 & 60.77 & 62.97 \\ \hline 
\multirow{2}{*}{A.6} & \multirow{2}{*}{CASIA} & \multirow{2}{*}{PGGAN} & \multirow{2}{*}{CASIA} & \multirow{2}{*}{PGGAN} & Original & 0.05 & 99.93 & 99.97 &4.62 & 95.37 & 95.40 &34.79 & 64.42 & 66.00  \\ 
 & & & & & GANprintR &8.37 & 93.64 & 93.66 & 27.77 & 72.21 & 72.22 & 39.15 & 60.02 & 61.70 \\ \hline

\end{tabular}
}
\end{table*}

\begin{figure}[!]
\centering
\begin{tikzpicture}
\node[anchor=north west] (A) at (0,0) {\includegraphics[width=\columnwidth]{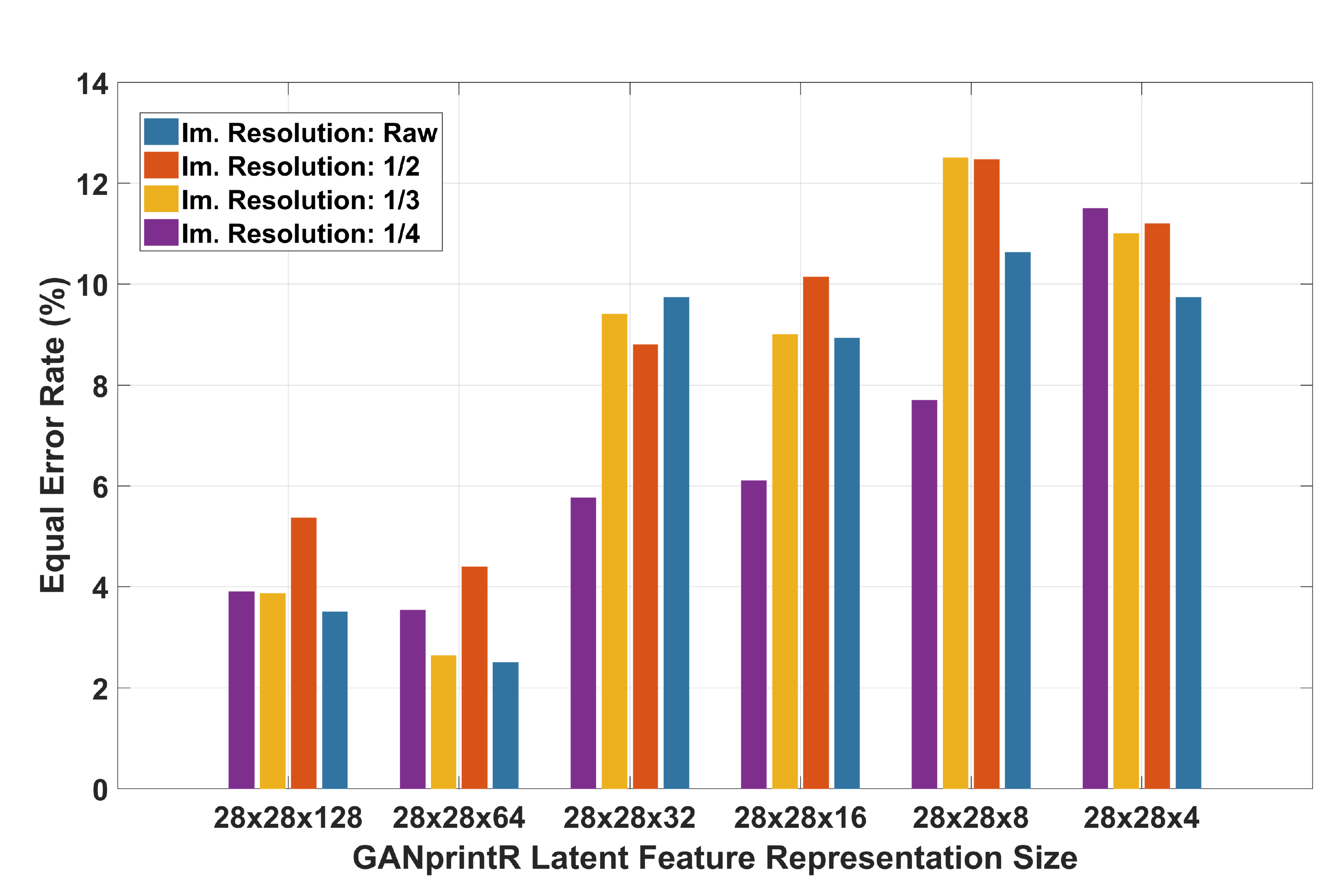}};
\end{tikzpicture}
\caption{\textbf{Robustness of the fake detection system trained with different resolutions and then tested with fakes improved with GANprintR under various configurations (representation sizes).} Five different GANprintR configurations are tested per image resolution level. The results observed point for the stability of  EER values with respect to using downsized synthetic images in training. This observation supports the conclusion that GANprintR is actually removing the GAN-fingerprint information.\label{fig5}}
\end{figure}

Finally, to confirm that GANprintR is actually removing the GAN-fingerprint information and not just reducing the image resolution of the images, we performed a final experiment where we trained the XceptionNet for fake detection considering different levels of image resolution, and then tested it using fakes improved with GANprintR. Fig.~\ref{fig5} shows the fake detection performance in terms of EER for different size of the latent feature representation of GANprintR. Five different GANprintR configurations are tested per image resolution. The obtained results point for the stability of EER values with respect to downsized synthetic images in training, concluding that our proposed approach is actually removing the GAN-fingerprint information.

\subsection{Impact of GANprintR on other Fake Detectors}\label{subsec:ganprintr_alldetectors}

For completeness, this section provides a comparative analysis of the impact of GANprintR on the three state-of-the-art fake detectors considered in this study. Table~\ref{tab:ganprintr_alldetectors} reports the EER and Recall observed when using the original fake images and the same ones after passing through GANprintR. 

In general, the same conclusions highlighted for XceptionNet in Sec.~\ref{subsec:UncontrainedScenrio} are extracted in Table~\ref{tab:ganprintr_alldetectors} for the other two fake detectors. For XceptionNet, an average absolute worsening of 9.65\% EER is produced when using GANprintR. This degradation is even higher for the Steganalysis fake detector with an average absolute worsening of 14.68\% EER. Finally, the fake detector based on Local Artifacts has proven to be the most robust one, with an average absolute worsening of 4.91\% EER. This lower performance degradation can be produced due to the higher EERs achieved in the original fake images (an average 35.54\% EER). These performance degradations prove the success of our proposed GANprintR, creating improved versions of the original fake images.

\section{Conclusions}\label{conclusions}
In this paper we presented a method (GANprintR) for improving the naturalness of facial fake images based on autoencoders, and we have empirically shown its ability to deceive state-of-the-art manipulation detection methods in a larger extent than some of the most sophisticate and realistic GAN-based synthetic face image generators available in the literature. Our method and experiments have been positioned and discussed in comparison with key related works around this problem published in the last couple of years. 

We started by training one deep autoencoder using public genuine face databases that models the typical spatial correlations between the pixels of real faces and simultaneously removes the high frequency components that correspond to the ``fingerprints" of the models used to generate synthetic images. In test time, the autoencoder was fed only with synthetic face images to produce manipulated versions, whose properties were deliberately changed for spoofing fake detection systems.

In the empirical validation of our approach, we used various well known face datasets, coming out with three major conclusions about the performance of the state-of-the-art fake detection methods: \textit{i)} the existing fake systems attain almost perfect performance when the evaluation data is derived from the same source used in the training phase, which suggests that these systems have actually learned the GAN ``fingerprints'' from the training fakes generated with GANs; \textit{ii)} the observed fake detection performance decreases substantially (over one order of magnitude) when the fake detection is exposed to data from unseen databases, and over seven times in case of substantially reduced image resolution; and \textit{iii)} the accuracy of the existing fake detection methods also drops significantly when analysing synthetic data manipulated by GANprintR.

In summary, our experiments suggest that the existing facial fake detection methods still have a poor generalisation capability and are highly susceptible to - even simple - image transformation manipulations, such as downsizing, image compression or others similar to the one proposed in this work. While loss of resolution may not be particularly concerning in terms of the potential misuse of the data, it is important to note that our approach is capable of confounding detection methods, while  maintaining a high visual similarity with the original image. 

Having shown some of the limitations of the state of the art in face manipulation detection, in future work we can harden such face manipulation detectors by exploiting our improved fakes. Additionally, further works may study: \emph{i)} how improved fakes obtained in similar ways as GANprintR can jeopardize other kinds of sensitive data (e.g., other popular biometrics like fingerprint~\cite{2019_TIFS_Fingerprint_PAs_Tolosana}, iris~\cite{Proenca_2019_CVPR_Workshops}, or behavioral traits \cite{2020_TIFS_BioTouchPass2_Tolosana,keystroke_fakeNews_2020}), \emph{ii)} how to improve the security of systems dealing with other kinds of sensitive data, and finally \emph{iii)} best ways to combine multiple manipulation detectors \cite{2018_INFFUS_MCSreview2_Fierrez} in a proper way to deal with the growing sophistication of fakes.

\section*{Acknowledgments}
\small This work has been supported by projects: PRIMA (H2020-MSCA-ITN-2019-860315), TRESPASS-ETN (H2020-MSCA-ITN-2019-860813), BIBECA (RTI2018-101248-B-I00 MINECO/FEDER), Bio-Guard (Ayudas Fundaci\'on BBVA a Equipos de Investigaci\'on Cient\'ifica 2017), Accenture, by NOVA LINCS (UIDB/04516/2020) with the financial support of FCT - Funda\c{c}\~ao para a Ci\^encia e a Tecnologia, through national funds, and by FCT/MCTES through national funds and co-funded by EU under the project UIDB/EEA/50008/2020.  We gratefully acknowledge the donation of the NVIDIA Titan X GPU used for this research made by NVIDIA Corporation. Ruben Tolosana is supported by Consejer\'ia de Educaci\'on, Juventud y Deporte de la Comunidad de Madrid y Fondo Social Europeo.



%

%

{
\bibliographystyle{IEEEtran}
\bibliography{egbib2}

\begin{thebibliography}{10}
\providecommand{\url}[1]{#1}
\csname url@samestyle\endcsname
\providecommand{\newblock}{\relax}
\providecommand{\bibinfo}[2]{#2}
\providecommand{\BIBentrySTDinterwordspacing}{\spaceskip=0pt\relax}
\providecommand{\BIBentryALTinterwordstretchfactor}{4}
\providecommand{\BIBentryALTinterwordspacing}{\spaceskip=\fontdimen2\font plus
\BIBentryALTinterwordstretchfactor\fontdimen3\font minus
  \fontdimen4\font\relax}
\providecommand{\BIBforeignlanguage}[2]{{%
\expandafter\ifx\csname l@#1\endcsname\relax
\typeout{** WARNING: IEEEtran.bst: No hyphenation pattern has been}%
\typeout{** loaded for the language `#1'. Using the pattern for}%
\typeout{** the default language instead.}%
\else
\language=\csname l@#1\endcsname
\fi
#2}}
\providecommand{\BIBdecl}{\relax}
\BIBdecl

\bibitem{BBCNews_deepfake}
\BIBentryALTinterwordspacing
{R. Cellan-Jones, Deepfake Videos Double in Nine Months}, 2019. [Online].
  Available: \url{https://www.bbc.com/news/technology-49961089}
\BIBentrySTDinterwordspacing

\bibitem{2015_ISPM_PAs}
A.~Hadid, N.~Evans, S.~Marcel, and J.~Fierrez, ``{Biometrics Systems Under
  Spoofing Attack: An Evaluation Methodology and Lessons Learned},'' \emph{IEEE
  Signal Processing Magazine}, vol.~32, no.~5, pp. 20--30, 2015.

\bibitem{Hernandez-Ortega2019}
J.~Hernandez-Ortega, J.~Fierrez, A.~Morales, and J.~Galbally, ``{Introduction
  to Face Presentation Attack Detection},'' in \emph{Handbook of Biometric
  Anti-Spoofing: Presentation Attack Detection}, S.~Marcel, M.~S. Nixon,
  J.~Fierrez, and N.~Evans, Eds.\hskip 1em plus 0.5em minus 0.4em\relax
  Springer International Publishing, 2019, pp. 187--206.

\bibitem{Galbally_Access_2014}
{Javier Galbally, Sebastien Marcel and Julian Fierrez}, ``{Biometric
  Anti-spoofing Methods: A Survey in Face Recognition},'' \emph{IEEE Access},
  vol.~2, pp. 1530--1552, 2014.

\bibitem{zao}
\BIBentryALTinterwordspacing
{ZAO}, 2019. [Online]. Available:
  \url{https://apps.apple.com/cn/app/id1465199127}
\BIBentrySTDinterwordspacing

\bibitem{faceapp}
\BIBentryALTinterwordspacing
{FaceApp}, 2017. [Online]. Available:
  \url{https://apps.apple.com/us/app/faceapp-ai-face-editor/id1180884341}
\BIBentrySTDinterwordspacing

\bibitem{Tolosana_2020_ARXIV}
R.~Tolosana, R.~Vera-Rodriguez, J.~Fierrez, A.~Morales, and J.~Ortega-Garcia,
  ``{DeepFakes and Beyond: A Survey of Face Manipulation and Fake Detection},''
  \emph{Information Fusion}, 2020.

\bibitem{Verdoliva_2020_arxiv}
L.~Verdoliva, ``{Media Forensics and DeepFakes: An Overview},''
  \emph{arXiv:2001.06564}, 2020.

\bibitem{Jain2019facialManipulation}
H.~Dang, F.~Liu, J.~Stehouwer, X.~Liu, and A.~Jain, ``{On the Detection of
  Digital Face Manipulation},'' in \emph{Proc. IEEE/CVF Conference on Computer
  Vision and Pattern Recognition}, 2020.

\bibitem{mccloskey2018detecting}
S.~McCloskey and M.~Albright, ``{Detecting GAN-Generated Imagery Using Color
  Cues},'' \emph{arXiv:1812.08247}, 2018.

\bibitem{yu2018attributing}
N.~Yu, L.~Davis, and M.~Fritz, ``{Attributing Fake Images to GANs: Analyzing
  Fingerprints in Generated Images},'' \emph{arXiv:1811.08180}, 2018.

\bibitem{Marra_MIPR_2018}
F.~{Marra}, D.~{Gragnaniello}, D.~{Cozzolino}, and L.~{Verdoliva}, ``{Detection
  of GAN-Generated Fake Images over Social Networks},'' in \emph{Proc. IEEE
  Conference on Multimedia Information Processing and Retrieval}, 2018.

\bibitem{wang2019fakespotter}
R.~Wang, L.~Ma, F.~Juefei-Xu, X.~Xie, J.~Wang, and Y.~Liu, ``{FakeSpotter: A
  Simple Baseline for Spotting AI-Synthesized Fake Faces},''
  \emph{arXiv:1909.06122}, 2019.

\bibitem{Yang_2019_ICASSP}
X.~{Yang}, Y.~{Li}, and S.~{Lyu}, ``{Exposing Deep Fakes Using Inconsistent
  Head Poses},'' in \emph{Proc. IEEE International Conference on Acoustics,
  Speech and Signal Processing}, 2019.

\bibitem{Matern_2019_WACVW}
F.~{Matern}, C.~{Riess}, and M.~{Stamminger}, ``Exploiting {V}isual {A}rtifacts
  to {E}xpose {D}eepfakes and {F}ace {M}anipulations,'' in \emph{Proc. IEEE
  Winter Applications of Computer Vision Workshops}, 2019.

\bibitem{He_2019_ICIP}
P.~{He}, H.~{Li}, and H.~{Wang}, ``Detection of {F}ake {I}mages {V}ia the
  {E}nsemble of {D}eep {R}epresentations from {M}ulti {C}olor {S}paces,'' in
  \emph{Proc. IEEE International Conference on Image Processing}, 2019.

\bibitem{wang2019detecting}
S.~Y. Wang, O.~Wang, A.~Owens, R.~Zhang, and A.~A. Efros, ``{Detecting
  Photoshopped Faces by Scripting Photoshop},'' in \emph{Proc. IEEE/CVF
  International Conference on Computer Vision}, 2019.

\bibitem{goodfellow2014generative}
I.~Goodfellow, J.~Pouget-Abadie, M.~Mirza, B.~Xu, D.~Warde-Farley, S.~Ozair,
  A.~Courville, and Y.~Bengio, ``{Generative Adversarial Nets},'' in
  \emph{Proc. Advances in Neural Information Processing Systems}, 2014.

\bibitem{NIST_challenge}
\BIBentryALTinterwordspacing
{National Institute of Standards and Technology. Media forensics challenge},
  2018. [Online]. Available:
  \url{https://www.nist.gov/itl/iad/mig/media-forensics-challenge-2018}
\BIBentrySTDinterwordspacing

\bibitem{vggface}
O.~Parkhi, A.~Vedaldi, and A.~Zisserman, ``{Deep Face Recognition},'' in
  \emph{Proc. British Machine Vision Conference}, 2015.

\bibitem{openface}
B.~Amos, B.~Ludwiczuk, and M.~Satyanarayanan, ``{OpenFace: A General-Purpose
  Face Recognition Library with Mobile Applications},'' \emph{CMU School of
  Computer Science}, 2016.

\bibitem{facenet}
F.~Schroff, D.~Kalenichenko, and J.~Philbin, ``{FaceNet: A Unified Embedding
  for Face Recognition and Clustering},'' in \emph{Proc. IEEE/CVF Conference on
  Computer Vision and Pattern Recognition}, 2015.

\bibitem{celebahq}
\BIBentryALTinterwordspacing
{CelebA-HQ}, 2018. [Online]. Available:
  \url{https://drive.google.com/drive/folders/0B4qLcYyJmiz0TXY1NG02bzZVRGs}
\BIBentrySTDinterwordspacing

\bibitem{Karras_2019_CVPR}
T.~Karras, S.~Laine, and T.~Aila, ``{A Style-Based Generator Architecture for
  Generative Adversarial Networks},'' in \emph{Proc. IEEE/CVF Conference on
  Computer Vision and Pattern Recognition}, 2019.

\bibitem{shen2019interpreting}
Y.~Shen, J.~Gu, X.~Tang, and B.~Zhou, ``{Interpreting the Latent Space of GANs
  for Semantic Face Editing},'' in \emph{Proc. IEEE/CVF Conference on Computer
  Vision and Pattern Recognition}, 2020.

\bibitem{celeba}
Z.~Liu, P.~Luo, X.~Wang, and X.~Tang, ``{Deep Learning Face Attributes in the
  Wild},'' in \emph{Proc. IEEE/CVF Int. Conf. on Computer Vision}, 2015.

\bibitem{rossler2019faceforensics++}
A.~R{\"o}ssler, D.~Cozzolino, L.~Verdoliva, C.~Riess, J.~Thies, and
  M.~Nie{\ss}ner, ``{FaceForensics++: Learning to Detect Manipulated Facial
  Images},'' in \emph{Proc. IEEE/CVF Int. Conf. on Computer Vision}, 2019.

\bibitem{pgan}
T.~Karras, T.~Aila, S.~Laine, and J.~Lehtinen, ``{Progressive Growing of GANs
  for Improved Quality, Stability, and Variation},'' in \emph{Proc.
  International Conference on Learning Representations}, 2018.

\bibitem{adobetool}
\BIBentryALTinterwordspacing
{Adjust and exaggerate facial features. Adobe Photoshop}, 2016. [Online].
  Available:
  \url{https://helpx.adobe.com/photoshop/how-to/face-aware-liquify.html}
\BIBentrySTDinterwordspacing

\bibitem{Albright_2019_CVPRW}
M.~Albright and S.~McCloskey, ``{Source Generator Attribution via Inversion?}''
  in \emph{Proc. IEEE/CVF Conference on Computer Vision and Pattern Recognition
  Workshops}, 2019.

\bibitem{Marra_2019_MIPR}
F.~Marra, D.~Gragnaniello, L.~Verdoliva, and G.~Poggi, ``{Do GANs Leave
  Artificial Fingerprints?}'' in \emph{Proc. IEEE Conference on Multimedia
  Information Processing and Retrieval}, 2019.

\bibitem{sngan}
T.~Miyato, T.~Kataoka, M.~Koyama, and Y.~Yoshida, ``{Spectral Normalization for
  Generative Adversarial Networks},'' in \emph{Proc. International Conference
  on Learning Representations}, 2018.

\bibitem{cramergan}
M.~Bellemare, I.~Danihelka, W.~Dabney, S.~Mohamed, B.~Lakshminarayanan,
  S.~Hoyer, and R.~Munos, ``{The Cramer Distance as a Solution to Biased
  Wasserstein Gradients},'' \emph{arXiv:1705.10743}, 2017.

\bibitem{mmdgan}
M.~Binkowski, D.~Sutherland, M.~Arbel, and A.~Gretton, ``{Demystifying MMD
  GANs},'' in \emph{Proc. International Conference on Learning
  Representations}, 2018.

\bibitem{zhang2019detecting}
X.~Zhang, S.~Karaman, and S.~Chang, ``{Detecting and Simulating Artifacts in
  GAN Fake Images},'' \emph{arXiv:1907.06515}, 2019.

\bibitem{huh2018fighting}
M.~Huh, A.~Liu, A.~Owens, and A.~A. Efros, ``{Fighting Fake News: Image Splice
  Detection Via Learned Self-Consistency},'' in \emph{Proc. European Conference
  on Computer Vision}, 2018.

\bibitem{zhou2018learning}
P.~Zhou, X.~Han, V.~I. Morariu, and L.~S. Davis, ``{Learning Rich Features for
  Image Manipulation Detection},'' in \emph{Proc. IEEE/CVF Conference on
  Computer Vision and Pattern Recognition}, 2018.

\bibitem{nataraj2019detecting}
L.~Nataraj, T.~Mohammed, B.~Manjunath, S.~Chandrasekaran, A.~Flenner, J.~Bappy,
  and A.~Roy-Chowdhury, ``{Detecting GAN Generated Fake Images Using
  Co-Occurrence Matrices},'' \emph{Electronic Imaging}, no.~5, pp. 1--7, 2019.

\bibitem{yi2014learning}
D.~Yi, Z.~Lei, S.~Liao, and S.~Li, ``{Learning Face Representation From
  Scratch},'' \emph{arXiv:1411.7923}, 2014.

\bibitem{cao2018vggface2}
Q.~Cao, , L.~Shen, W.~Xie, O.~Parkhi, and A.~Zisserman, ``{VGGFace2: A Dataset
  for Recognising Faces Across Pose and Age},'' in \emph{Proc. IEEE Int. Conf.
  on Automatic Face \& Gesture Recognition}, 2018.

\bibitem{FFHQ}
\BIBentryALTinterwordspacing
{Flickr-Faces-HQ Dataset (FFHQ)}, 2019. [Online]. Available:
  \url{https://github.com/NVlabs/ffhq-dataset}
\BIBentrySTDinterwordspacing

\bibitem{100kfaces}
\BIBentryALTinterwordspacing
{100,000 Faces Generated by AI}, 2018. [Online]. Available:
  \url{https://generated.photos/}
\BIBentrySTDinterwordspacing

\bibitem{Kazemi_2014_CVPR}
V.~Kazemi and J.~Sullivan, ``{One {M}illisecond {F}ace {A}lignment with an
  {E}nsemble of {R}egression {T}rees},'' in \emph{Proc. IEEE/CVF Conference on
  Computer Vision and Pattern Recognition}, 2014.

\bibitem{chollet2017xception}
F.~Chollet, ``{Xception: Deep Learning with Depthwise Separable
  Convolutions},'' in \emph{Proc. IEEE/CVF Conference on Computer Vision and
  Pattern Recognition}, 2017.

\bibitem{dolhansky2019deepfake}
B.~Dolhansky, R.~Howes, B.~Pflaum, N.~Baram, and C.~C. Ferrer, ``{The Deepfake
  Detection Challenge (DFDC) Preview Dataset},'' \emph{arXiv:1910.08854}, 2019.

\bibitem{deng2009imagenet}
J.~Deng, W.~Dong, R.~Socher, L.~J. Li, K.~Li, and L.~Fei-Fei, ``{ImageNet: A
  Large-Scale Hierarchical Image Database},'' in \emph{Proc. IEEE/CVF
  Conference on Computer Vision and Pattern Recognition}, 2009.

\bibitem{guo2016msceleb}
Y.~Guo, L.~Zhang, Y.~Hu, X.~He, and J.~Gao, ``M{S}-{C}eleb-1{M}: {A} {D}ataset
  and {B}enchmark for {L}arge {S}cale {F}ace {R}ecognition,'' in \emph{Proc.
  European Conference on Computer Vision}, 2016.

\bibitem{korshunov2018deepfakes}
P.~Korshunov and S.~Marcel, ``{DeepFakes: a New Threat to Face Recognition?
  Assessment and Detection},'' \emph{arXiv:1812.08685}, 2018.

\bibitem{zhu2017unpaired}
J.~Y. Zhu, T.~Park, P.~Isola, and A.~A. Efros, ``{Unpaired Image-to-Image
  Translation Using Cycle-Consistent Adversarial Networks},'' in \emph{Proc.
  IEEE/CVF International Conference on Computer Vision}, 2017.

\bibitem{2019_TIFS_Fingerprint_PAs_Tolosana}
R.~Tolosana, M.~Gomez-Barrero, C.~Busch, and J.~Ortega-Garcia, ``{Biometric
  Presentation Attack Detection: Beyond the Visible Spectrum},'' \emph{IEEE
  Transactions on Information Forensics and Security}, 2019.

\bibitem{Proenca_2019_CVPR_Workshops}
H.~Proenca and J.~C. Neves, ``{Segmentation-Less and Non-Holistic Deep-Learning
  Frameworks for Iris Recognition},'' in \emph{Proc. IEEE/CVF Conf. on Computer
  Vision and Pattern Recognition Workshops}, 2019.

\bibitem{2020_TIFS_BioTouchPass2_Tolosana}
R.~Tolosana, R.~Vera-Rodriguez, J.~Fierrez, and J.~Ortega-Garcia,
  ``{BioTouchPass2: Touchscreen Password Biometrics Using Time-Aligned
  Recurrent Neural Networks},'' \emph{IEEE Transactions on Information
  Forensics and Security}, vol.~15, pp. 2616--2628, 2020.

\bibitem{keystroke_fakeNews_2020}
A.~Morales, A.~Acien, J.~Fierrez, J.~V. Monaco, R.~Tolosana, R.~Vera-Rodriguez,
  and J.~Ortega-Garcia, ``{Keystroke Biometrics in Response to Fake News
  Propagation in a Global Pandemic},'' in \emph{Proc. IEEE Computer Software
  and Applications Conference Workshops}, 2020.

\bibitem{2018_INFFUS_MCSreview2_Fierrez}
J.~Fierrez, A.~Morales, R.~Vera-Rodriguez, and D.~Camacho, ``{Multiple
  Classifiers in Biometrics. Part 2: Trends and Challenges},''
  \emph{Information Fusion}, vol.~44, pp. 103--112, 2018.

\end{thebibliography}
}

\begin{IEEEbiography}[{\includegraphics[width=1in,height=1.25in,clip,keepaspectratio]{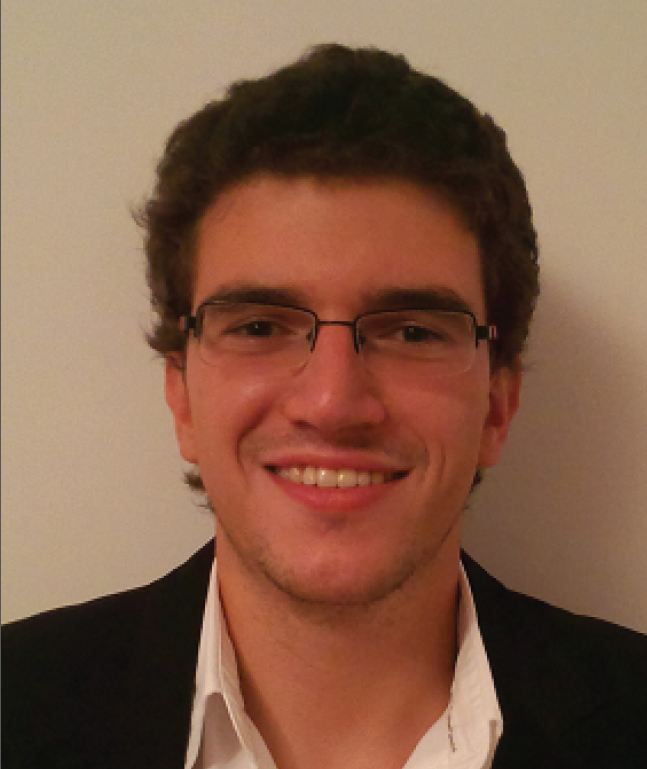}}]%
{Jo\~{a}o Neves} received the B.Sc., M.Sc., and Ph.D degrees in Computer Science from the University of Beira Interior, Portugal, in 2011, 2013, and 2018, respectively. He is currently an assistant professor at University of Beira Interior. His research interests broadly include computer vision and pattern recognition, with a particular focus on biometrics and surveillance. He is author of several publications and also collaborates as a reviewer in many different high-impact conferences (e.g., WAVC, IJCB, ACM MM, etc.) and journals (e.g., IEEE TMI, TIFS, TCYB, TCSVT, etc.).
\end{IEEEbiography}

\begin{IEEEbiography}[{\includegraphics[width=1in,height=1.25in,clip,keepaspectratio]{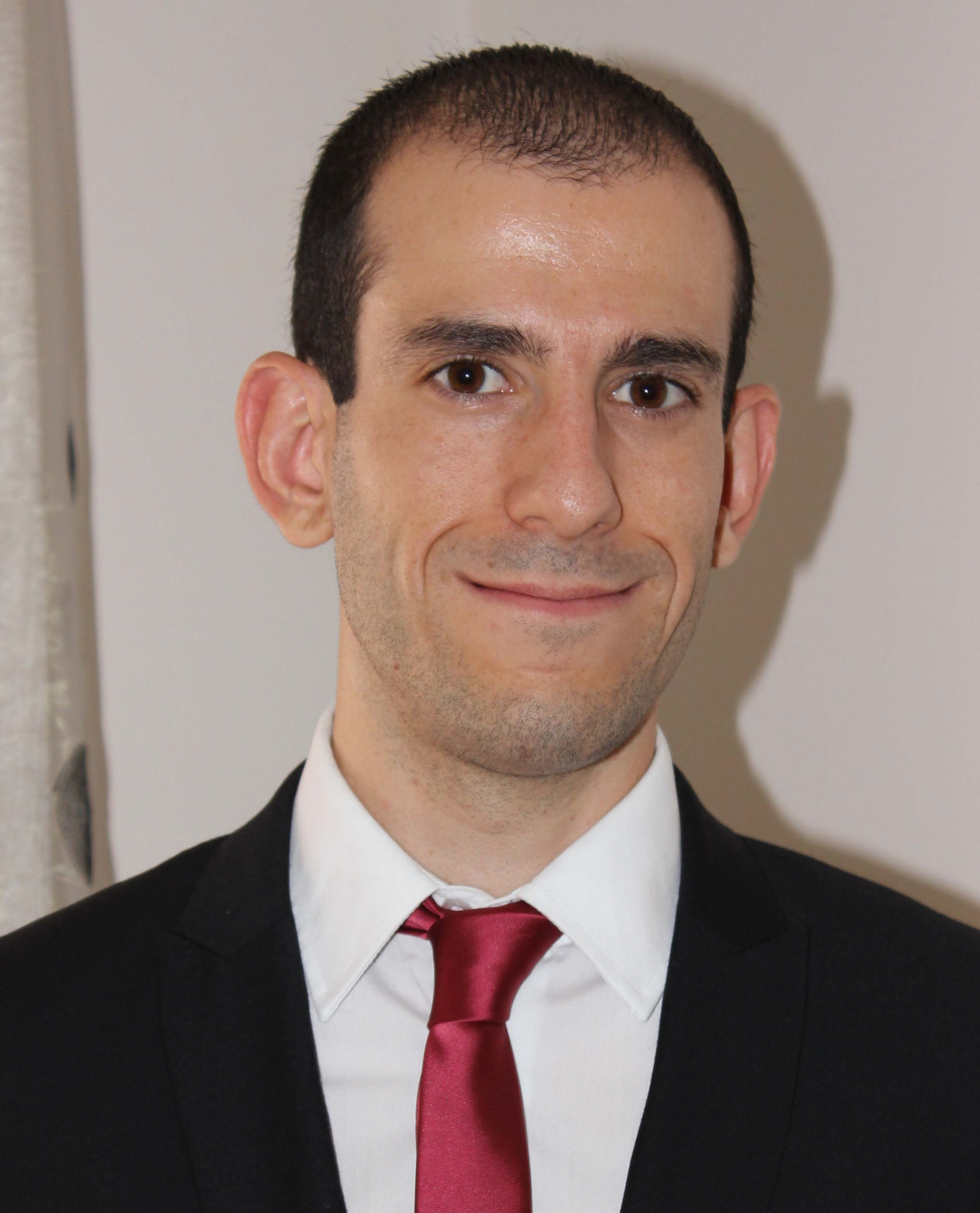}}]%
{Ruben Tolosana}
received the M.Sc. degree in Telecommunication Engineering, and his Ph.D. degree in Computer and Telecommunication Engineering, from Universidad Autonoma de Madrid, in 2014 and 2019, respectively. In April 2014, he joined the Biometrics and Data Pattern Analytics - BiDA Lab at the Universidad Autonoma de Madrid, where he is currently collaborating as a PostDoctoral researcher. Since then, Ruben has been granted with several awards such as the FPU research fellowship from Spanish MECD (2015), and the European Biometrics Industry Award (2018). His research interests are mainly focused on signal and image processing, pattern recognition, deep learning, and biometrics, particularly in the areas of handwriting and handwritten signature. He is author of several publications and also collaborates as a reviewer in many different high-impact conferences (e.g., ICDAR, ICB, BTAS, EUSIPCO, etc.) and journals (e.g., IEEE TPAMI, TIFS, TCYB, TIP, ACM Computing Surveys, etc.). Finally, he has participated in several National and European projects focused on the deployment of biometric security through the world.
\end{IEEEbiography}

\begin{IEEEbiography}[{\includegraphics[width=1in,height=1.25in,clip,keepaspectratio]{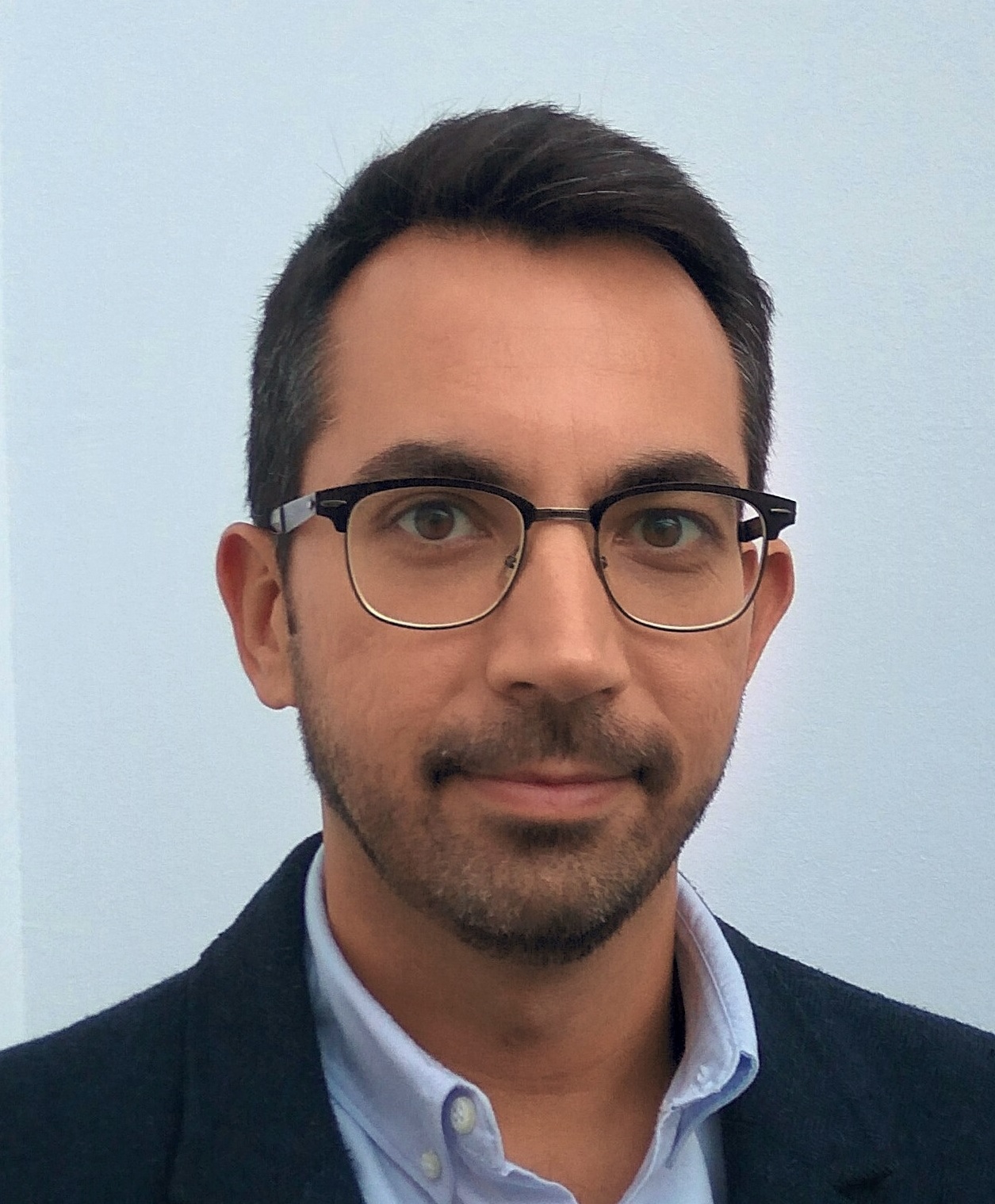}}]{Ruben Vera-Rodriguez} received the M.Sc. degree in telecommunications engineering from Universidad de Sevilla, Spain, in 2006, and the Ph.D. degree in electrical and electronic engineering from Swansea University, U.K., in 2010. Since 2010, he has been affiliated with the Biometric Recognition Group, Universidad Autonoma de Madrid, Spain, where he is currently an Associate Professor since 2018. His research interests include signal and image processing, pattern recognition, and biometrics, with emphasis on signature, face, gait verification and forensic applications of biometrics. He is actively involved in several National and European projects focused on biometrics. Ruben has been Program Chair for the IEEE 51st International Carnahan Conference on Security and Technology (ICCST) in 2017; and the 23rd Iberoamerican Congress on Pattern Recognition (CIARP 2018) in 2018.
\end{IEEEbiography}

\begin{IEEEbiography}[{\includegraphics[width=1in,height=1.25in,clip,keepaspectratio]{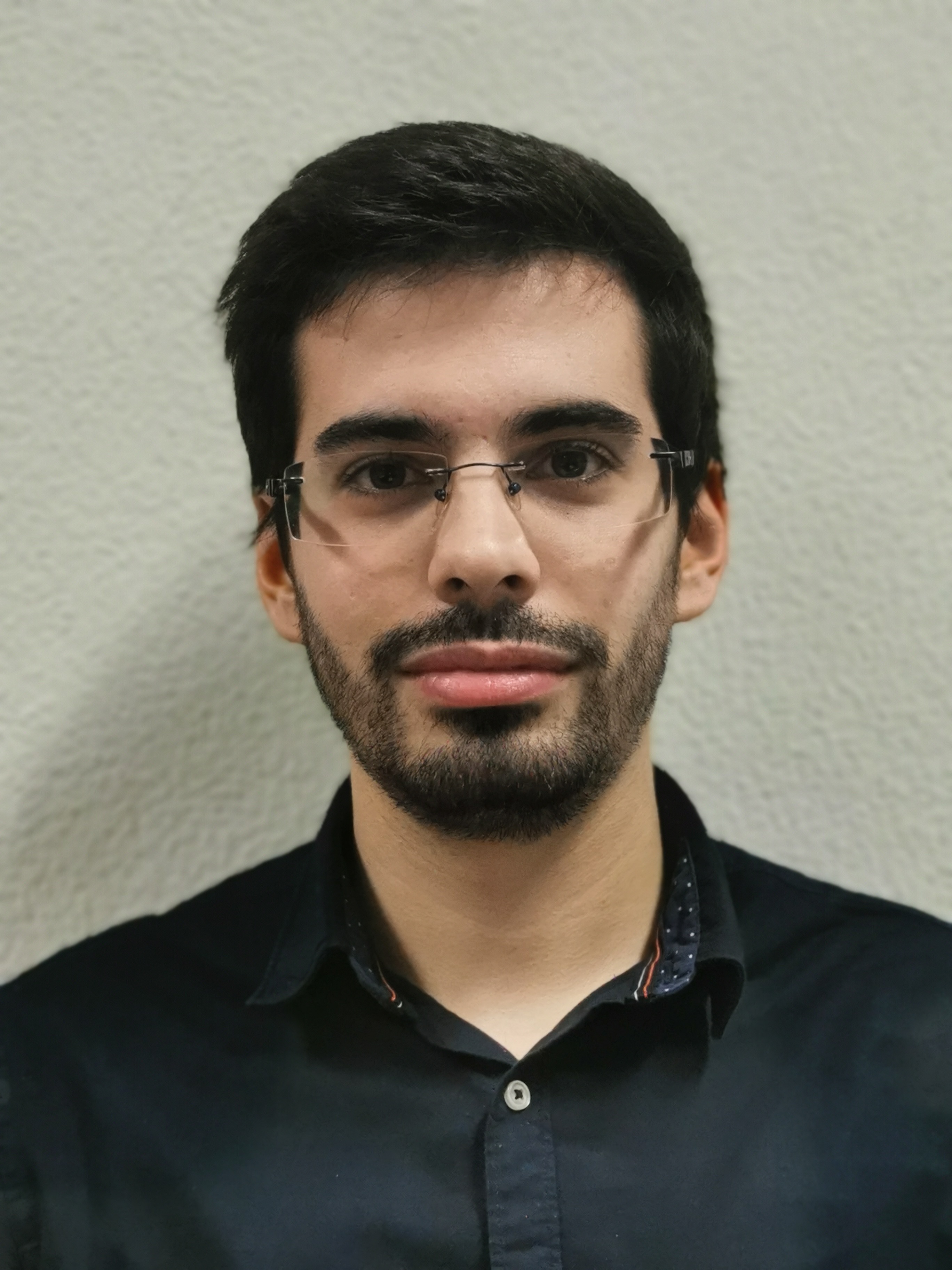}}]%
{Vasco Lopes}, received his BSc (2017) and MSc (2019) degrees in Computer Science and Engineering from the University of Beira Interior, Portugal, where he is currently pursing a PhD in the field of Artificial Intelligence, with focus on computer vision. His current research interests broadly include computer vision, robotics and artificial intelligence. He was the recipient of the APRP Best Dissertation in Pattern Recognition 2019 Award.
\end{IEEEbiography}

\begin{IEEEbiography}[{\includegraphics[width=1in,height=1.25in,clip,keepaspectratio]{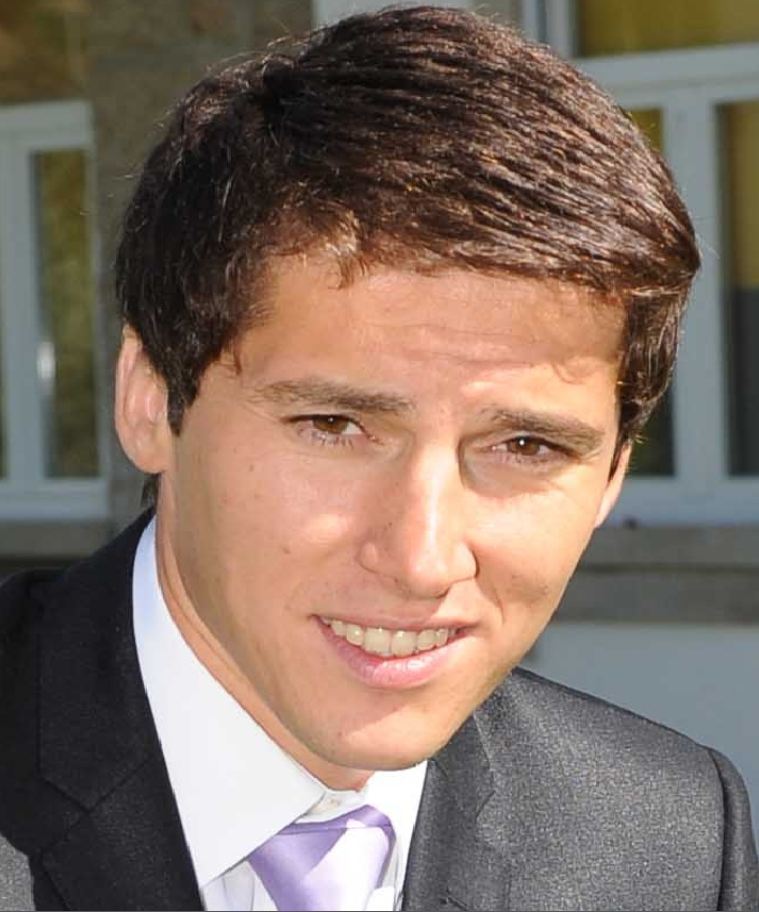}}]%
{Hugo Proen\c{c}a} B.Sc. (2001), M.Sc. (2004) and Ph.D. (2007) is an Associate Professor in the Department of Computer Science, University of Beira Interior and has been researching mainly about biometrics and visual-surveillance. He is the coordinating editor of the IEEE Biometrics Council Newsletter and the area editor (ocular biometrics) of the IEEE Biometrics Compendium Journal. He is a member of the Editorial Boards of the Image and Vision Computing and International Journal of Biometrics and served as Guest  Editor of special issues of the Pattern Recognition Letters, Image and Vision Computing and Signal, Image and Video Processing journals.
\end{IEEEbiography}

\begin{IEEEbiography}
[{\includegraphics[width=1in,height=1.25in,clip,keepaspectratio]{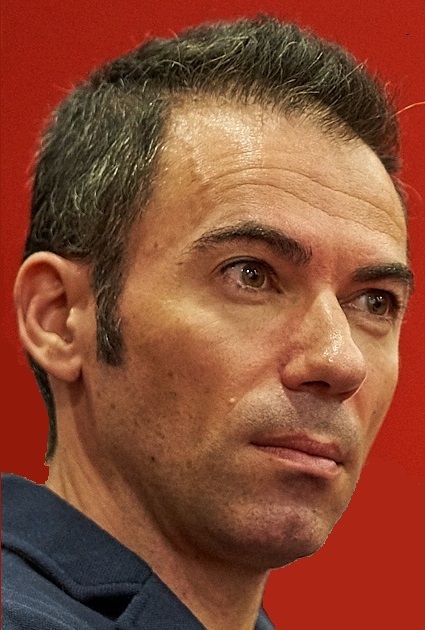}}]%
{Julian Fierrez}{\,}received the M.Sc. and Ph.D. degrees in telecommunications engineering from the Universidad Politecnica de Madrid, Spain, in 2001 and 2006, respectively. Since 2004 he has been at Universidad Autonoma de Madrid, where he is currently an Associate Professor. From 2007 to 2009 he was a Visiting Researcher with Michigan State University, USA, under a Marie Curie postdoc. His research is on signal and image processing, HCI, responsible AI, and biometrics for security and human behavior analysis. He is actively involved in large EU projects in these topics (e.g., TABULA RASA and BEAT in the past, now IDEA-FAST and TRESPASS-ETN), and has attracted notable impact for his research. He was a recipient of a number of distinctions, including the EAB Industry Award 2006, the EURASIP Best Ph.D. Award 2012, and the 2017 IAPR Young Biometrics Investigator Award. He has received best paper awards at ICB and ICPR. He is Associate Editor of the IEEE TRANSACTIONS ON INFORMATION FORENSICS AND SECURITY and the IEEE TRANSACTIONS ON IMAGE PROCESSING. He is member of the ELLIS Society.
\end{IEEEbiography}

\end{document}